\documentclass[runningheads]{llncs}

\usepackage[T1]{fontenc}
\usepackage{graphicx}
\usepackage[table]{xcolor}
\usepackage{hyperref}
\usepackage{url}
\usepackage{orcidlink}
\usepackage{booktabs}
\usepackage{array}
\usepackage{amsmath}
\usepackage{makecell}
\usepackage{multirow}

\usepackage{amsmath, amssymb} 
\usepackage{pgfplots}
\usepgfplotslibrary{groupplots}
\usepgfplotslibrary{statistics} 
\pgfplotsset{compat=1.18}

\definecolor{lightgray}{gray}{0.9}

\usepackage{orcidlink}

\makeatletter
\newcommand{\printfnsymbol}[1]{%
  \textsuperscript{\@fnsymbol{#1}}%
}
\makeatother
\begin{document}
\title{GEAR: Generative Editing for Scene-Conditioned Outfit Retrieval for VR Travel}

\titlerunning{GEAR: Generative Editing for Scene-Conditioned Outfit Retrieval for VR Travel}

\author{
Gia-Huy Dinh\inst{1,2}\orcidlink{}\thanks{These authors contributed equally to this work.} \and
Thinh-Phuc Nguyen\inst{1, 2}\orcidlink{} \printfnsymbol{1}  \and
Thanh-Hai Nguyen\inst{1, 2}\orcidlink{} \printfnsymbol{1}  \and
Lam-Huy Nguyen-Ngoc\inst{1,2}\orcidlink{} \and
Hoang-Bach Nguyen-Phan\inst{1,2}\orcidlink{} \and
Nam-Quan Nguyen\inst{1,2}\orcidlink{} \and
Minh-Triet Tran\inst{1,2}\orcidlink{0000-0003-3046-3041} \and
Trung-Nghia Le\inst{1,2}\orcidlink{0000-0002-7363-2610}\thanks{Corresponding author.} 
}

\authorrunning{Gia-Huy Dinh et al.}

\institute{
University of Science, Ho Chi Minh City, Vietnam
\and
Vietnam National University, Ho Chi Minh City, Vietnam
\\
\email{
\{dhghuy23, nltphuc23, nthai23, nnlhuy23, nphbach23, nnquan23\}@apcs.fitus.edu.vn, \\
\{tmtriet, ltnghia\}@fit.hcmus.edu.vn
}
}

\maketitle
\begin{abstract}
In virtual environments, choosing appropriate outfits requires contextual harmony with surrounding scenes, yet bridging the visual gap between 3D ambient views and catalog garment assets remains challenging. This paper investigates scene-conditioned outfit retrieval for interactive 3D environments, where a user's egocentric view serves as the retrieval signal. We present GEAR (Generative Editing for Asset Retrieval), a framework that bridges the cross-domain gap by synthesizing intermediate visual queries rather than predicting direct compatibility. GEAR uses a vision-language model to describe suitable clothing from a captured scene, employs an instruction-guided diffusion model to render the concept onto a standardized reference character, and utilizes the cropped clothing region to retrieve existing 2D outfit assets via embedding similarity. An interactive Unity prototype demonstrates real-time scene capture, asset retrieval, and in-engine 3D avatar fitting. We benchmark GEAR against direct visual similarity, aesthetic scoring, and VLM-based reranking on a controlled dataset using automated Swiss-system pairwise judgments, where GEAR achieves competitive top-ranked performance (mean Elo@5 of 1012.75 and top-10 overlap of 0.180) alongside 0.539 pairwise agreement. Furthermore, a user study across 20 virtual travel scenes demonstrates that GEAR achieves the highest overall participant preference in both visual appeal (34.5\%) and contextual suitability (34.5\%).
\end{abstract}

\section{Introduction}

Selecting appropriate clothing is inherently contextual: an outfit that appears visually appealing in isolation may become unsuitable once the surrounding environment, activity, and occasion are taken into account. Traditional fashion recommendation and retrieval systems have largely focused on 2D product catalogs and user queries expressed through text, reference images, or historical preferences. Immersive applications, including virtual reality (VR) and desktop 3D environments, motivate using the current scene view as a query for this contextual recommendation task. An egocentric view offers one way to capture the user's surroundings without requiring a manually specified outfit query.

\begin{figure*}[t]
    \centering
    \includegraphics[width=\textwidth]{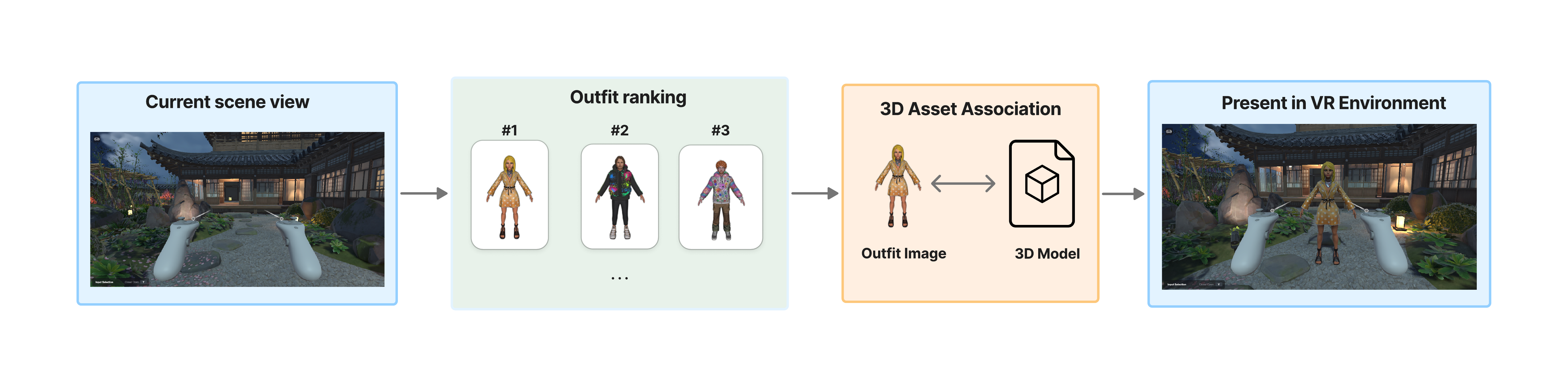}
    \caption{Overview of scene-conditioned outfit retrieval in an
interactive 3D environment. The user's current scene view
serves as a query to retrieve and rank outfits according to
their contextual compatibility. A selected outfit is matched
to its corresponding 3D asset, when available, and presented
on an avatar in the original Unity environment.}
    \label{fig:problem_setting}
\end{figure*}

This paper studies scene-conditioned fashion retrieval in interactive
3D environments, where the user's current view serves as a query for
retrieving suitable outfit assets. As illustrated in
Figure~\ref{fig:problem_setting}, the system captures an egocentric
scene view and returns ranked outfit recommendations from an existing
asset collection. Retrieved outfits are presented as 2D previews,
with corresponding 3D models displayed in Unity when available. A retrieval-based formulation preserves existing asset fidelity while
avoiding the challenges of directly generating and rigging 3D garments.
By representing outfits as standardized 2D images, retrieval can be
performed independently of 3D asset presentation, with corresponding
models instantiated in the virtual environment upon selection.

To bridge the semantic gap between scene views and fashion assets,
we propose GEAR (Generative Editing for Asset Retrieval), which
constructs scene-conditioned visual queries through image generation.
A vision--language model (VLM) generates an outfit description
from the scene, which an instruction-guided diffusion model
renders onto a reference character. The resulting clothing
region serves as a visual query for retrieving existing outfit
assets. This approach is intended to reduce the visual mismatch between
scenery and garment images without generating new assets
as retrieval outputs. To demonstrate this retrieval mechanism in context, we implement an interactive 
Unity-based prototype paired with a curated fashion asset collection derived from 
Ready Player Me. The catalog includes 1,624 standardized 2D outfit renderings 
used for indexing, alongside 105 corresponding 3D models available for 
avatar presentation in the virtual environment.

We evaluate GEAR against three baseline retrieval paradigms: CLIP-style direct 
visual similarity, composite aesthetic scoring, and composed VLM retrieval with 
reranking. On a controlled benchmark of 10 scenes and 100 candidate outfits, we 
evaluate complete rankings using an automated Swiss-system pairwise tournament. In addition, we conduct a user study with 10 participants
across 20 travel scenes to assess the perceived visual appeal
and contextual suitability of top-three recommendation sets.

In summary, this work makes the following contributions:
\begin{itemize}
    \item We study scene-conditioned retrieval of existing outfit assets
    in a virtual-environment application, using a scene view as the query and
    linking 2D retrieval results to available 3D assets.
    \item We propose GEAR, a retrieval pipeline that uses VLM reasoning and 
    instruction-guided character editing to synthesize intermediate visual 
    queries for searching existing asset collections.
    \item We implement an end-to-end Unity prototype backed by 1,624 outfit images,
    with 105 corresponding 3D models, connecting scene capture to ranked
    previews and optional in-engine avatar presentation.
    \item We benchmark GEAR against direct visual, aesthetic, and composed VLM 
    retrieval baselines through both automated pairwise preferences and human 
    evaluations of visual appeal and context fit.
\end{itemize}

\section{Related Work}

\subsection{Scene-Conditioned Fashion Retrieval}

Fashion retrieval has incorporated language and contextual information beyond garment appearance. Fashion IQ studies retrieval using reference images and natural-language feedback \cite{wu2021fashion}. Scene-aware approaches instead model the suitability of clothing for a given environment. SAFRS learns scene--outfit compatibility by combining scene and outfit representations \cite{10.1145/3543507.3583435}, while GLOSS additionally considers body shape and relational information \cite{11175554}. Unlike these approaches, GEAR constructs a visual query from the user's current view of an interactive 3D environment to retrieve existing outfit assets. Our evaluation compares GEAR with direct visual retrieval,
aesthetic scoring, and composed VLM retrieval
(Section~\ref{sec:baselines}). The benchmark lacks the body-shape
annotations and user--outfit relationship graphs required by the released
GLOSS pipeline. Direct comparisons with SAFRS
and GLOSS remain a limitation and an important direction for further evaluation.

\subsection{Generation-Guided Retrieval}

Generative models can construct intermediate visual representations for retrieving existing items. In fashion, GeCo generates complementary garment templates from seed items through image-to-image translation and incorporates the generated images into composed retrieval queries \cite{attimonelli2024fashion}. Beyond fashion, Imagine and Seek generates an imagined
proxy conditioned on a reference image and relative text to support composed image retrieval~\cite{li2025imagine}.
GEAR applies generation-guided retrieval to scene-conditioned outfit recommendation: an environmental scene guides an outfit
description, which is rendered on a standardized reference
character to construct the visual retrieval query.

\subsection{Instruction-Guided Image Editing}

Instruction-guided image editing enables visual modifications specified through natural-language prompts. InstructPix2Pix demonstrates this capability using a conditional diffusion model \cite{brooks2023instructpix2pix}. Related virtual try-on methods, such as IDM-VTON, instead condition generation on an existing garment image to visualize that garment on a person \cite{choi2024improving}. GEAR uses image editing for a different purpose: synthesizing a scene-conditioned outfit appearance without requiring a reference garment. The edited image is subsequently cropped and embedded for similarity-based retrieval, separating visual query construction from the selection of existing fashion assets.

\section{Methodology}

\subsection{Overview}

\begin{figure}[t]
    \centering
    \includegraphics[width=\linewidth]{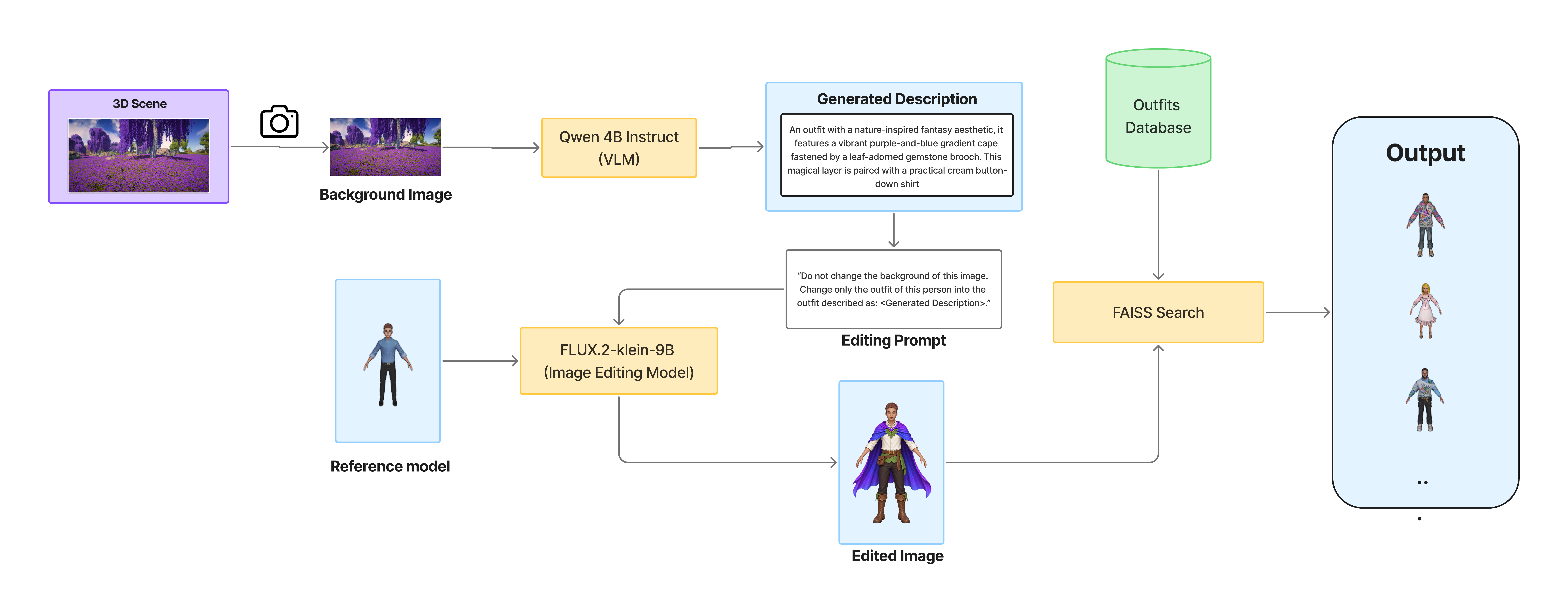}
    \caption{GEAR retrieval pipeline. A VLM suggests clothing for the
    scene, and FLUX edits a reference character to wear it. The clothing region
    of the generated image is cropped and used to search an indexed outfit
    collection. The generated image is an intermediate visual query; the
    returned items are existing outfit assets.}
    \label{fig:image_edit_method}
\end{figure}

Our system retrieves existing outfit assets for a user's current view of an
interactive 3D environment. GEAR (Generative Editing for Asset Retrieval)
constructs a visual outfit query from that view and searches for similar
assets. Figure~\ref{fig:image_edit_method} shows the sequence from scene
understanding to image editing and retrieval.

\subsection{Problem Formulation}

Let $I_s$ denote an egocentric scene image and $\mathcal{C}$ a collection of
candidate outfit assets represented by images $I_c$, $c\in\mathcal{C}$. The
objective is to rank existing outfit assets according to their suitability
for the observed scene. Rather than directly predicting scene--outfit
compatibility, GEAR constructs a scene-conditioned visual query and ranks
candidates by their appearance similarity to this representation.

The retrieved item is an existing asset, not a generated garment. When a
candidate has an associated local \texttt{.glb} model, that model can be
presented in Unity; image-only candidates remain valid retrieval results.

\subsection{Proposed Method}

GEAR uses a reference character to provide a consistent canvas for rendering
the outfit concept. The intended benefit is to express clothing appearance
in the same visual modality as the indexed candidates. Whether this improves
scene suitability depends on both the generated concept and its match to the
available collection.

\paragraph{Scene-conditioned outfit suggestion.}

A vision--language model examines $I_s$ and generates a textual description
$t_s$ of an outfit suited to its setting. The description supplies visual
details for the subsequent edit. Optional user preferences or feedback can
also inform the suggestion, but neither was provided in the reported
benchmark. This VLM produces the outfit description for GEAR; the
separate VLM Retrieval baseline uses a different retrieval pipeline.

\paragraph{Reference-character image editing.}

We provide a reference image $I_r$ of a clothed virtual character. The system
contains male and female reference characters and selects one according to
the request; the benchmark uses the worker's default male reference. An
instruction-guided image editor receives $I_r$ and $t_s$ and is asked to
change only the character's outfit:
\begin{equation}
    I_e = \operatorname{Edit}(I_r, t_s).
\end{equation}
The generated image $I_e$ depicts one possible scene-conditioned look. We
apply a fixed spatial crop, $I_q=\operatorname{CropClothes}(I_e)$, intended to
reduce the influence of the face and surrounding background. This heuristic
assumes a centered, front-facing character and does not perform garment
segmentation. Image editing constructs $I_q$, but no generated garment is
inserted into the retrieval database. The fixed crop is intended to focus retrieval on clothing,
but it may retain body or background cues or exclude garment details.
Sensitivity to reference-character appearance and framing has not
been evaluated.

\paragraph{Outfit retrieval.}

We encode the cropped query $I_q$ and each candidate outfit image $I_c$
with the same PE-Core image encoder $f$. Their $L_2$-normalized embeddings
are compared by inner product:
\begin{equation}
    s(c \mid I_s) =
    \frac{f(I_q)}{\lVert f(I_q)\rVert_2}^{\!\top}
    \frac{f(I_c)}{\lVert f(I_c)\rVert_2}.
\end{equation}
A FAISS index stores the candidate embeddings~\cite{douze2025faiss}, and the candidates are
returned in descending order of $s(c \mid I_s)$. This last step maps the
generated visual target to outfits that already exist in the collection.
The score measures appearance similarity to the synthesized target; its
relationship to scene suitability is assessed by the retrieval evaluation.
For the benchmark, retrieval is restricted to the 100 candidates in the
shared manifest and returns a complete ranking of that pool.

\subsection{Unity Prototype}

We integrate GEAR into a Unity prototype~\cite{unity} intended for
virtual travel applications. The system uses the user's egocentric view as the retrieval query and presents ranked recommendations through an in-engine gallery. Users can inspect the retrieved outfits and, when corresponding 3D assets are available, visualize them on an avatar within the original environment.

This integration separates scene-conditioned retrieval from asset presentation, allowing the same ranking framework to accommodate both image-only candidates and outfits with associated 3D models. The experimental evaluation focuses on retrieval quality rather than end-to-end VR interaction performance. The reported experiments use static scene images and recommendation
sets; in-headset interaction remain unevaluated.

\section{Experiments}

\subsection{Dataset}

Our asset collection comprises 1,624 outfit images from Ready Player
Me~\cite{readyplayerme}, including 105 outfits with corresponding
\texttt{.glb} models. Outfit images are rendered using a standardized
front-facing camera with transparent backgrounds for retrieval.
Assets with available 3D models can additionally be presented in Unity.

For evaluation, we constructed a fixed subset of 10 scene image
queries and 100 candidate outfits sampled from our asset collection.
All methods are evaluated on the same candidate pool to ensure
a controlled comparison of scene-conditioned retrieval performance.

\subsection{Experimental Setup}
Each method ranks the same 100 candidate outfits for each of the
10 scene queries, producing complete and directly comparable rankings.

Following pairwise multimodal evaluation approaches used in fashion
try-on and text-to-3D generation~\cite{kwon2026feat,wu2024gpt},
we construct automated reference preferences using a blinded
Swiss-style tournament judged by Gemma 4 26B A4B~\cite{team2026gemma}.
For each scene, seven rounds pair every outfit once per round,
without repeating pairs, yielding 350 comparisons per scene
and 3,500 distinct comparisons overall.

Each pair is evaluated in both presentation orders to reduce
positional bias, yielding 7,000 presentations.
The judge considers climate, season, activity, occasion,
style, theme, color harmony, and overall suitability.
A preference is retained only when both orders agree;
otherwise, the comparison is treated as a tie.

We aggregate these judgments into per-scene reference rankings
using Elo ratings initialized at 1,000 with $K=32$.
These automated rankings serve as evaluation references,
not independently validated human ground truth. The 3,500 comparisons are drawn from 10 shared scene queries. Presenting both orders addresses positional
consistency but does not validate the judge's suitability judgments.

\subsection{Baseline Methods}
\label{sec:baselines}
\subsubsection{CLIP-based Retrieval}
The baseline labeled CLIP uses the PE-Core-L14-336 image encoder \cite{bolya2026perception} to embed
the scene and each outfit directly. Candidates are ranked by cosine
similarity. This implements a CLIP-style visual retrieval strategy \cite{radford2021learning} with
PE-Core weights, without an explicit outfit suggestion or generated query.

\subsubsection{Aesthetic Predictor}
This baseline evaluates the visual appeal of an outfit within a given scene. For each candidate outfit, the outfit image is composited onto the query scene background. The resulting image is passed through a CLIP image encoder \cite{cherti2023reproducible} followed by a learned linear aesthetic prediction head \cite{schuhmann2022laionaesthetics}, which outputs a scalar aesthetic score. Outfits are ranked according to the predicted aesthetic quality of the composed image.

\subsubsection{VLM Retrieval}
The VLM-based baseline uses Qwen3-VL-4B-Instruct \cite{bai2025qwen3} to describe the scene and propose several
suitable and unsuitable outfits. It combines scene and outfit descriptions
with visual scene features to form multiple PE-Core queries, retrieves
candidates with FAISS, and reranks their images with
Qwen3-VL-Reranker-2B \cite{li2026qwen3}. This is a separate retrieval baseline; the VLM in
GEAR supplies only the outfit suggestion for the edit. This baseline differs from GEAR in query construction,
encoder configuration, and reranking; it therefore compares
complete pipelines rather than isolating image generation.

\subsection{Evaluation Metrics}

Our reference ranking follows the pattern of pairwise VLM judgments
aggregated with Elo ratings, introduced for text-to-3D evaluation by
Wu et al.~\cite{wu2024gpt} and adapted to fashion try-on by
Kwon et al.~\cite{kwon2026feat}. We adapt that pattern to scene--outfit
suitability; the metrics below evaluate how closely each retrieval method
matches our reference judgments.

We use three metrics that assess
both pairwise ranking agreement and the quality of top-ranked
recommendations.
\paragraph{Pairwise Agreement.}
We measure agreement of the ranked outfits with the automated
reference judgments. For each judged pair, a method receives
$1$ if its ranking agrees with the reference preference,
$0$ if the preference is reversed, and $0.5$ if the
reference judgment is tied. Scores are averaged across
all judged pairs within each scene. Under this scoring rule, a uniformly random ranking
has expected pairwise agreement of $0.5$, with tied reference
judgments contributing $0.5$ regardless of the ranking.

\paragraph{Mean Elo@5.}
We compute the average reference Elo rating of a method's
top-five retrieved outfits. This metric measures the
reference quality of the recommended items independently
of their exact ranking order.

\paragraph{Top-10 Overlap.}
We measure the proportion of outfits in the reference
top-10 that also appear in the method's top-10:
\begin{equation}
    \mathrm{Overlap}@10 =
    \frac{|T_{10} \cap E_{10}|}{10},
\end{equation}
where $T_{10}$ and $E_{10}$ denote the method's and
reference ranking's top-10 sets, respectively.

All metrics are macro-averaged across the 10 benchmark scenes.
\subsection{Implementation Details}

The four methods ranked the same 100 candidates.
Candidate embeddings were precomputed, $L_2$-normalized, and searched with
flat FAISS inner-product indices. Direct retrieval and GEAR shared a
PE-Core-L14-336 image index; VLM Retrieval used a separate PE-Core-B16-224
image--text index.

For GEAR, Qwen3-VL-4B-Instruct generated an outfit description of at most
50 words from an RGB scene image resized to a maximum side length of
512 pixels (temperature $0.7$, top-$p$ $0.9$, 256-token limit).
FLUX.2-klein-9B~\cite{flux-2-2025} edited the default male reference character at
$1024\times1024$ pixels using 30 steps, guidance scale $7.0$;
the reference image was resized to $512\times512$ pixels. The query crop
retained the central 50\% of image width and lower 65\% of image height.

The aesthetic baseline used an OpenCLIP ViT-L/14 encoder with a pretrained
LAION aesthetic head. VLM Retrieval generated seven compatible and three
incompatible outfit descriptions, reranked all 100 candidates with
Qwen3-VL-Reranker-2B. No optional user
preferences were supplied. The interactive prototype ran in Unity
6000.2.1f1~\cite{unity} on Meta Quest 2 with Meta XR SDK v78.


\begin{table}[t]
    \centering
    \caption{Results on the retrieval benchmark with 10 scene queries
    and 100 outfit candidates. Values are macro-averages over scenes; higher is
    better for every metric. Pairwise agreement was the benchmark's declared
    primary metric; mean Elo@5 and top-10 overlap characterize the quality
    of top-ranked recommendations. The best value in each metric column is shown in bold.}
    \label{tab:retrieval_benchmark}
    \small
    \begin{tabular}{lccc}
        \toprule
        \textbf{Method}
        & \shortstack{\textbf{Pairwise}\\\textbf{agreement}}
        & \shortstack{\textbf{Mean}\\\textbf{Elo@5}}
        & \shortstack{\textbf{Top-10}\\\textbf{overlap}} \\
        \midrule
        CLIP & 0.506 & 998.80 & 0.140 \\
        Aesthetic Predictor & 0.480 & 995.60 & 0.080 \\
        VLM & \textbf{0.548} & 1012.59 & 0.170 \\
        \textbf{GEAR (ours)} & 0.539 & \textbf{1012.75}
        & \textbf{0.180} \\
        \bottomrule
    \end{tabular}
\end{table}

\subsection{Quantitative Results}

Table~\ref{tab:retrieval_benchmark} compares the four retrieval
methods. VLM achieves the highest pairwise agreement (0.548),
followed by GEAR (0.539), CLIP (0.506), and Aesthetic Predictor
(0.480).

GEAR has the highest observed mean Elo@5 (1012.75 versus 1012.59
for VLM) and top-10 overlap (0.180 versus 0.170), and the absolute differences
are 0.16 Elo points and 0.010 overlap.  VLM achieves higher
pairwise agreement. The statistical reliability of
these benchmark differences has not been assessed.

\newcommand{\qualimg}[1]{%
    \includegraphics[
        width=\linewidth,
        height=1.6cm,
        keepaspectratio
    ]{images/qualitative/#1.png}%
}

\newcommand{\qualrow}[7]{%
    #1 &
    \qualimg{#2} &
    \qualimg{#3} &
    \qualimg{#4} &
    \qualimg{#5} &
    \qualimg{#6} &
    \qualimg{#7}\\
    \addlinespace[4pt]
}

\begin{figure*}[t]
    \centering
    \small
    \setlength{\tabcolsep}{3pt}
    \renewcommand{\arraystretch}{1.15}

    \begin{tabular}{
        @{}
        >{\raggedright\arraybackslash}p{0.13\textwidth}
        *{3}{>{\centering\arraybackslash}p{0.125\textwidth}}
        !{\hspace{5pt}\color{gray!50}\vrule width 0.5pt\hspace{5pt}}
        *{3}{>{\centering\arraybackslash}p{0.125\textwidth}}
        @{}
    }

        &
        \multicolumn{3}{c}{
            \includegraphics[
                width=0.27\textwidth,
                height=2.7cm,
                keepaspectratio
            ]{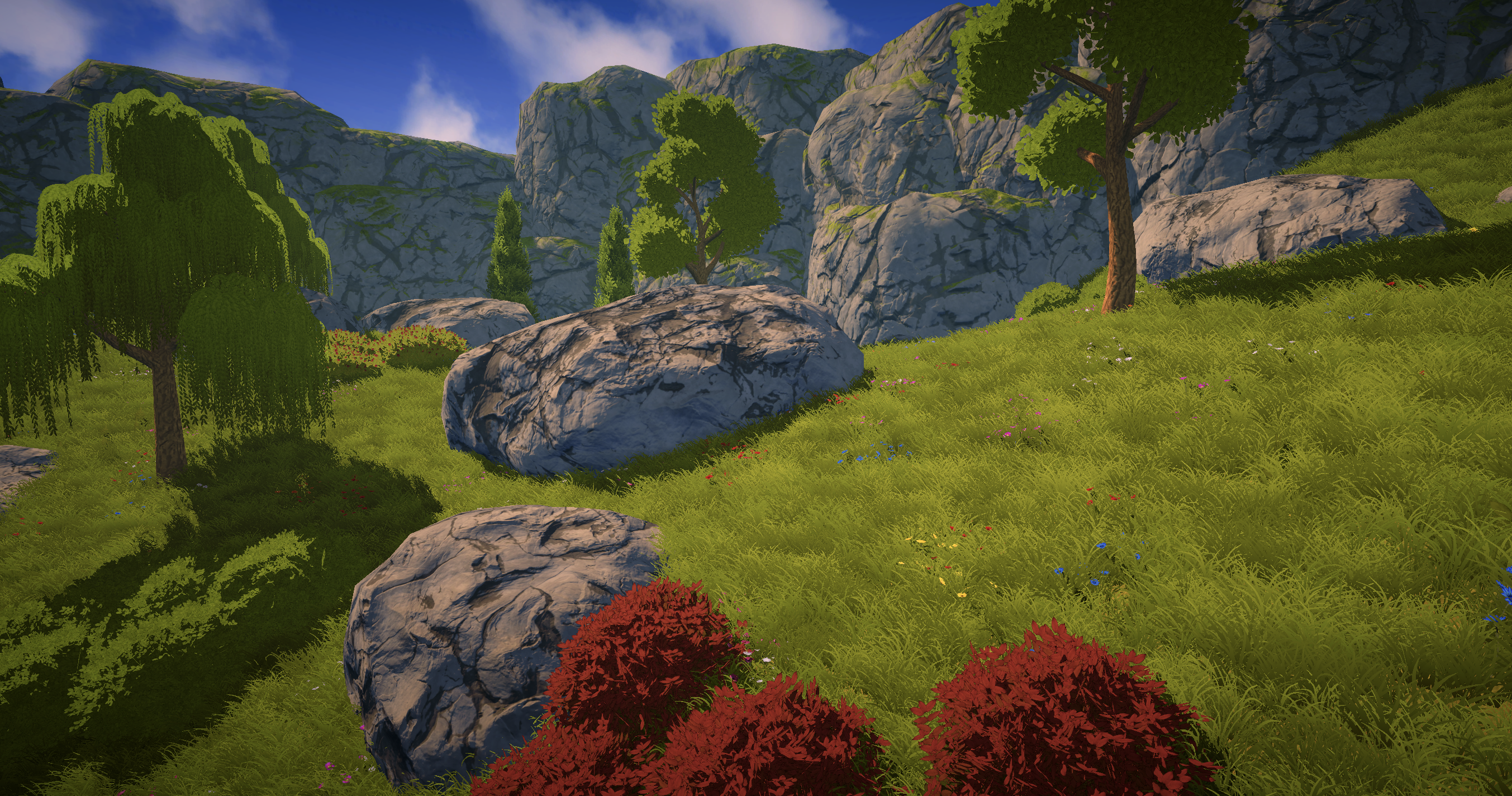}
        }
        &
        \multicolumn{3}{c}{
            \includegraphics[
                width=0.27\textwidth,
                height=2.7cm,
                keepaspectratio
            ]{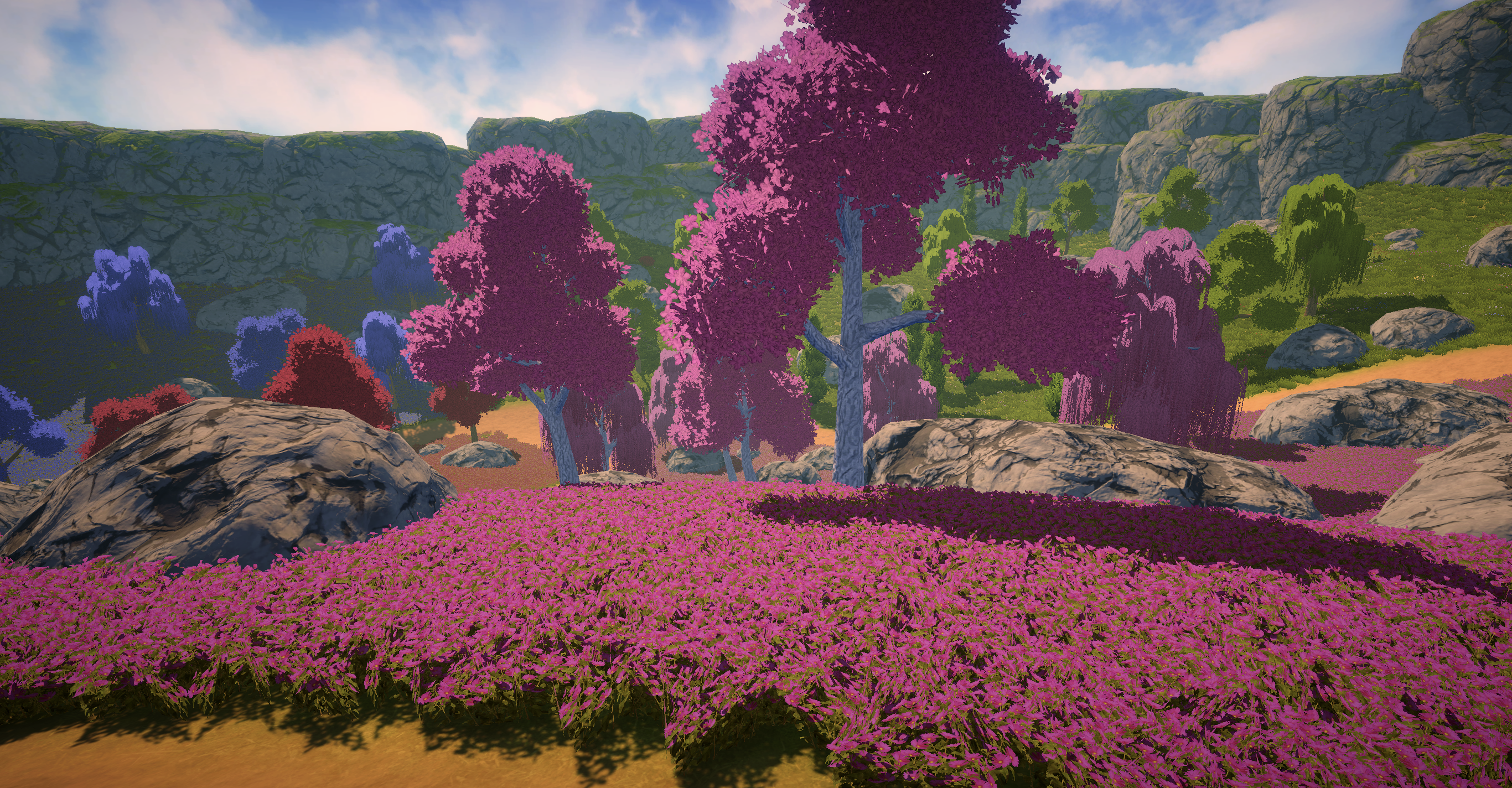}
        }\\[6pt]

        \toprule
        \textbf{Method}
        & \textbf{Top-1}
        & \textbf{Top-2}
        & \textbf{Top-3}
        & \textbf{Top-1}
        & \textbf{Top-2}
        & \textbf{Top-3}\\

        \midrule

        \qualrow{CLIP}
            {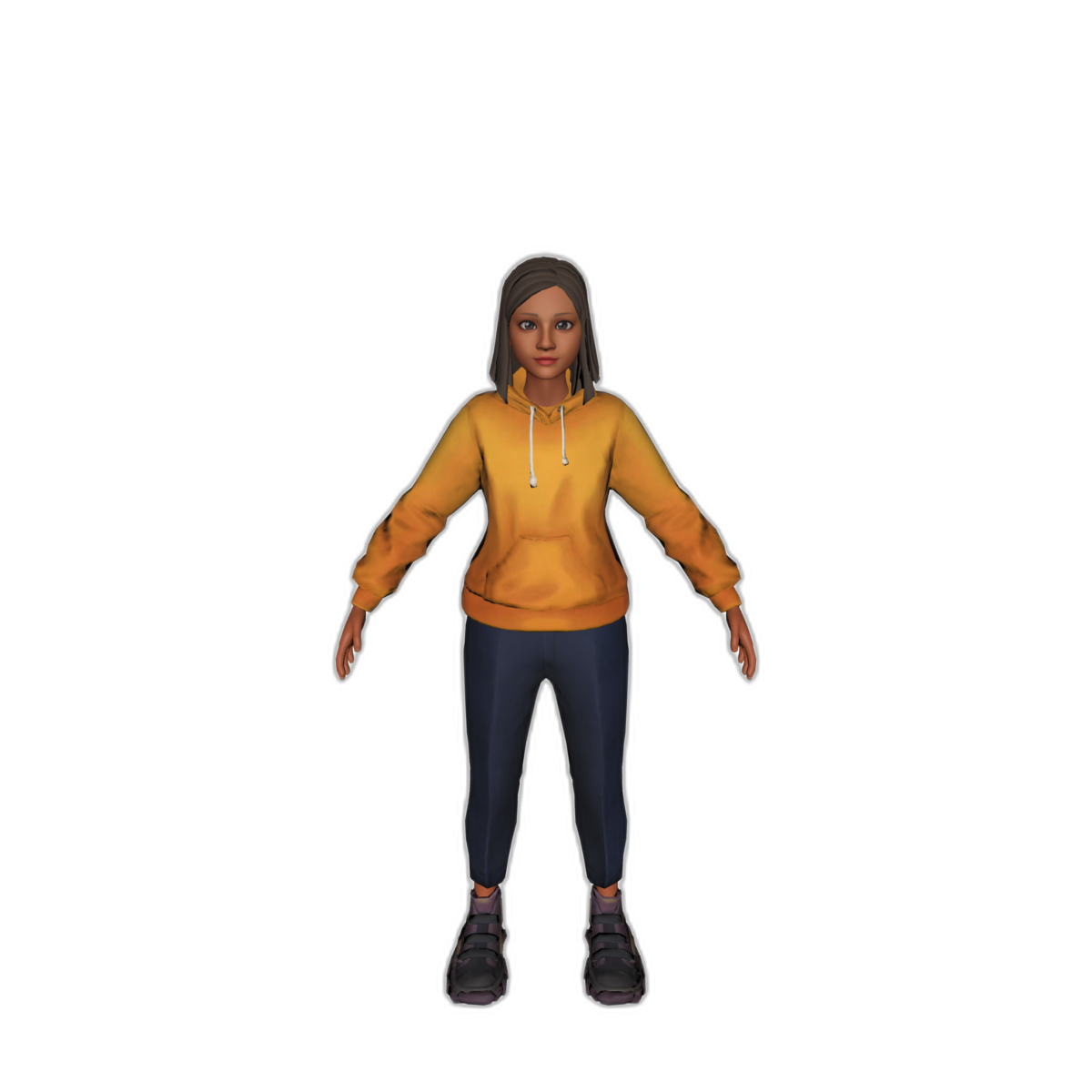}{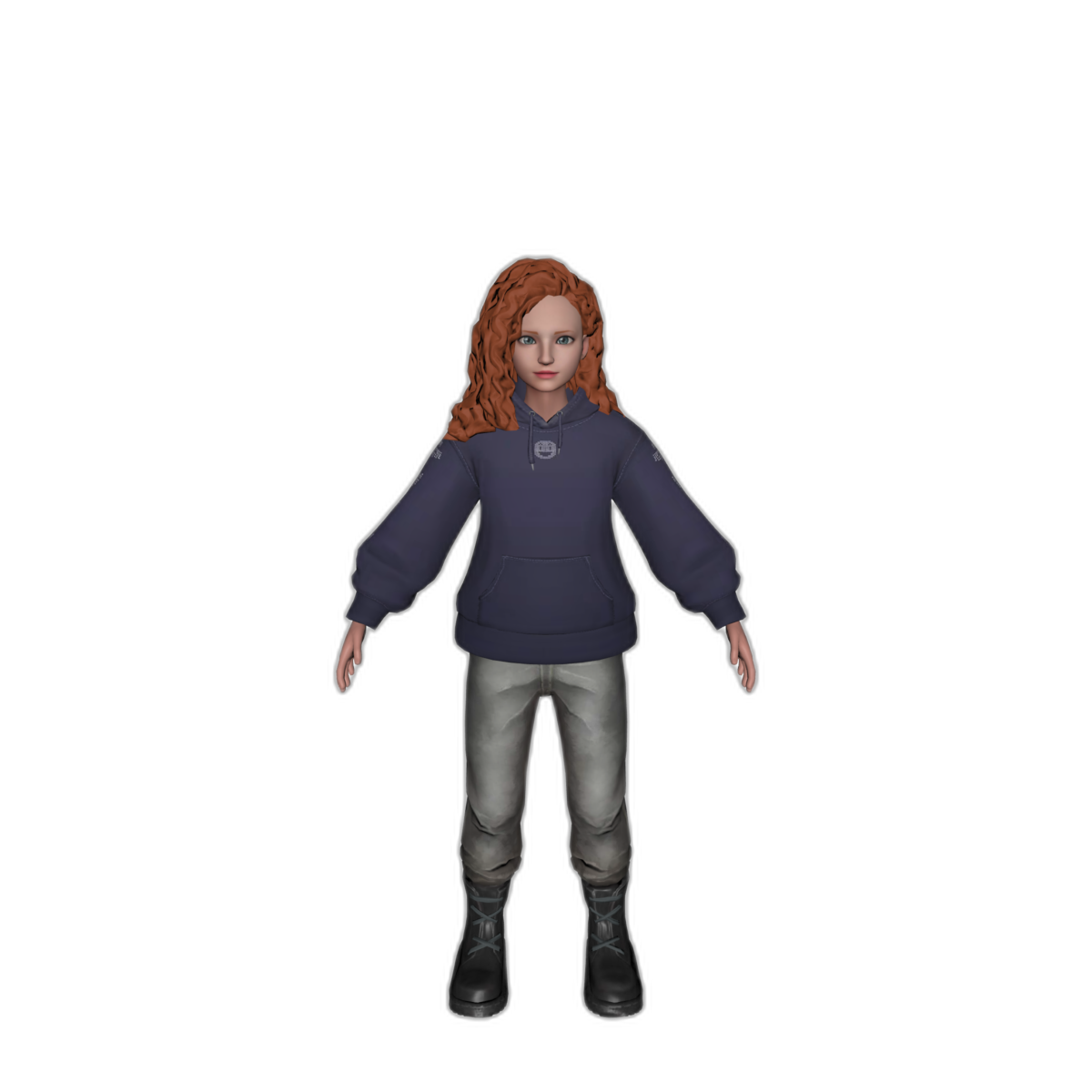}{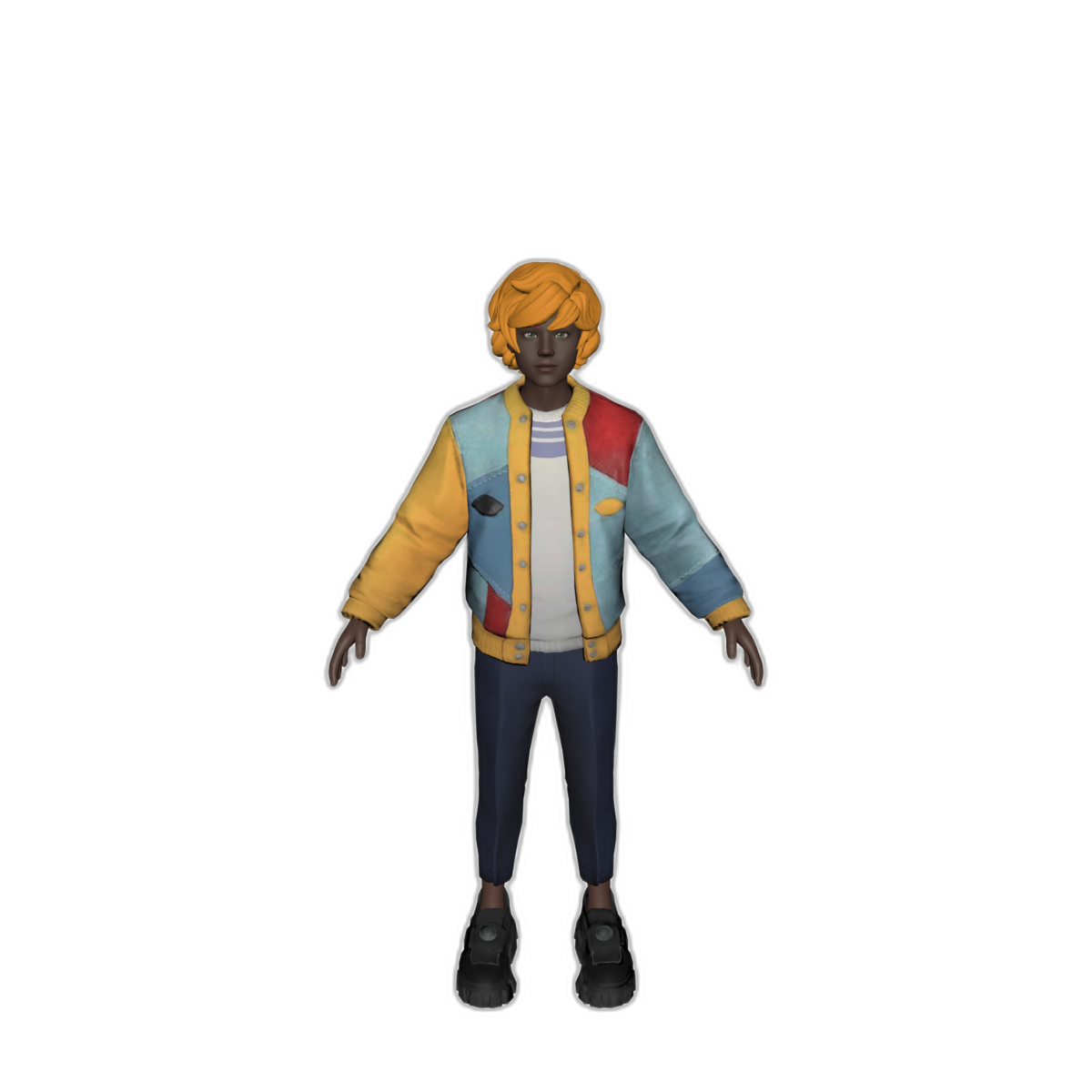}
            {1}{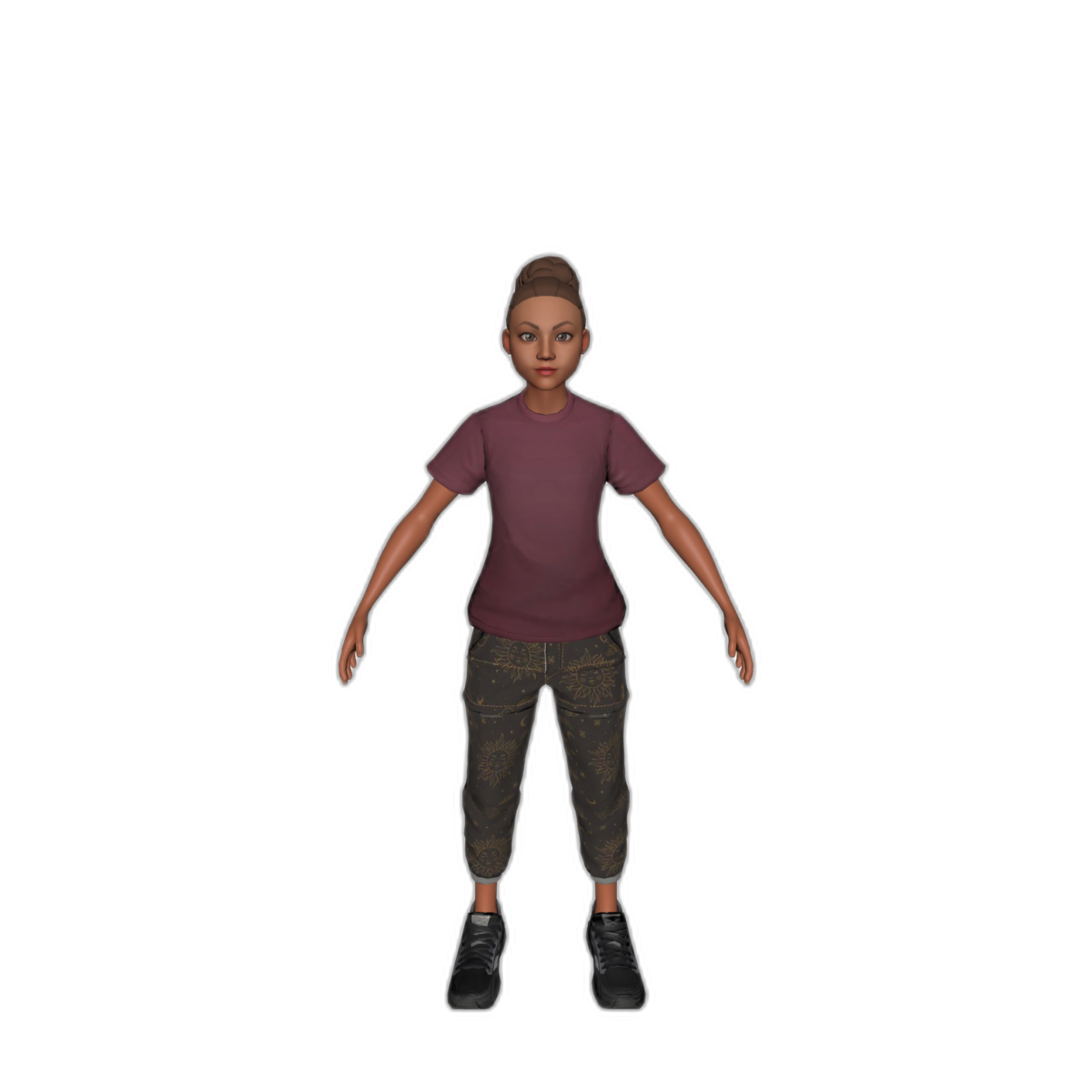}{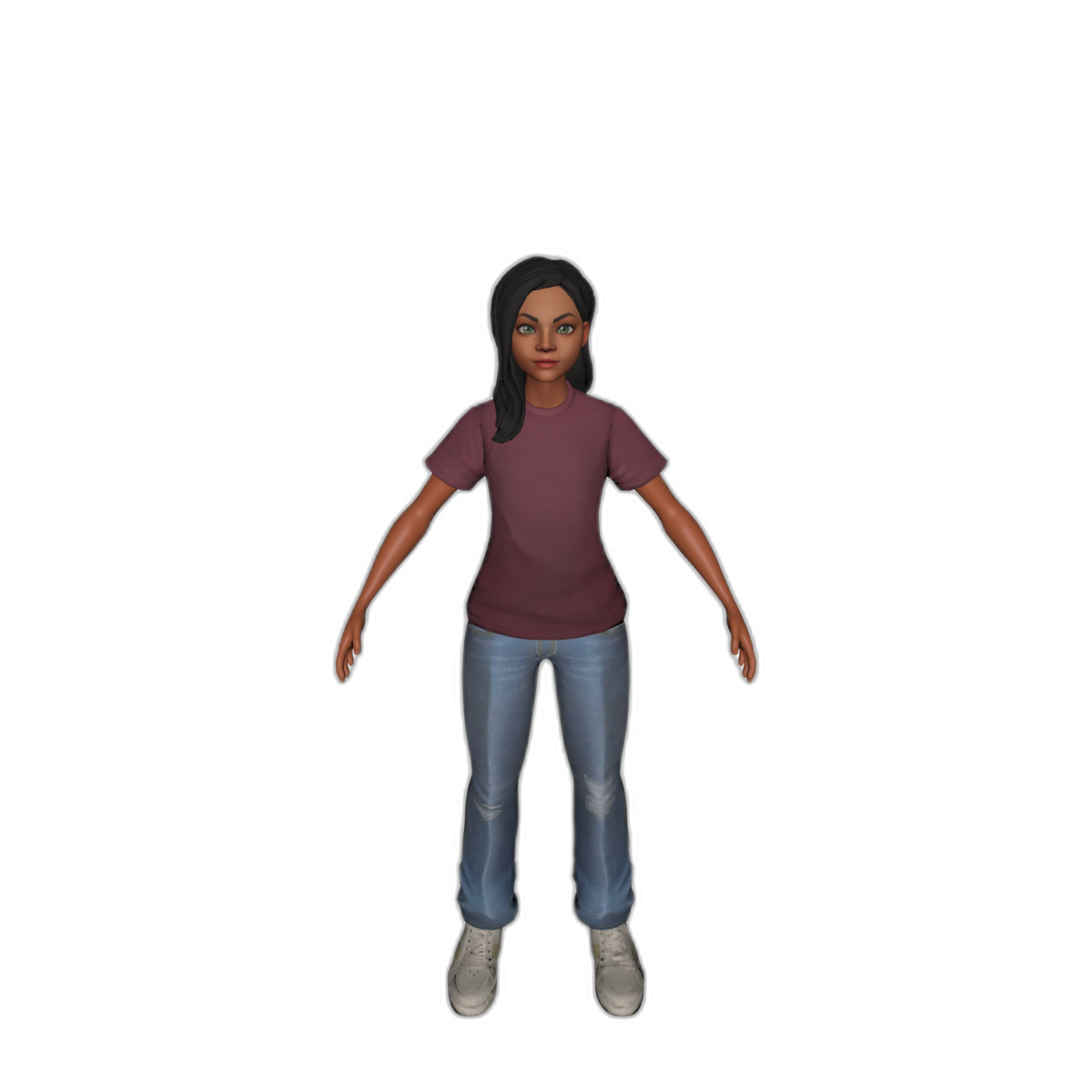}

        \qualrow{Aesthetic}
            {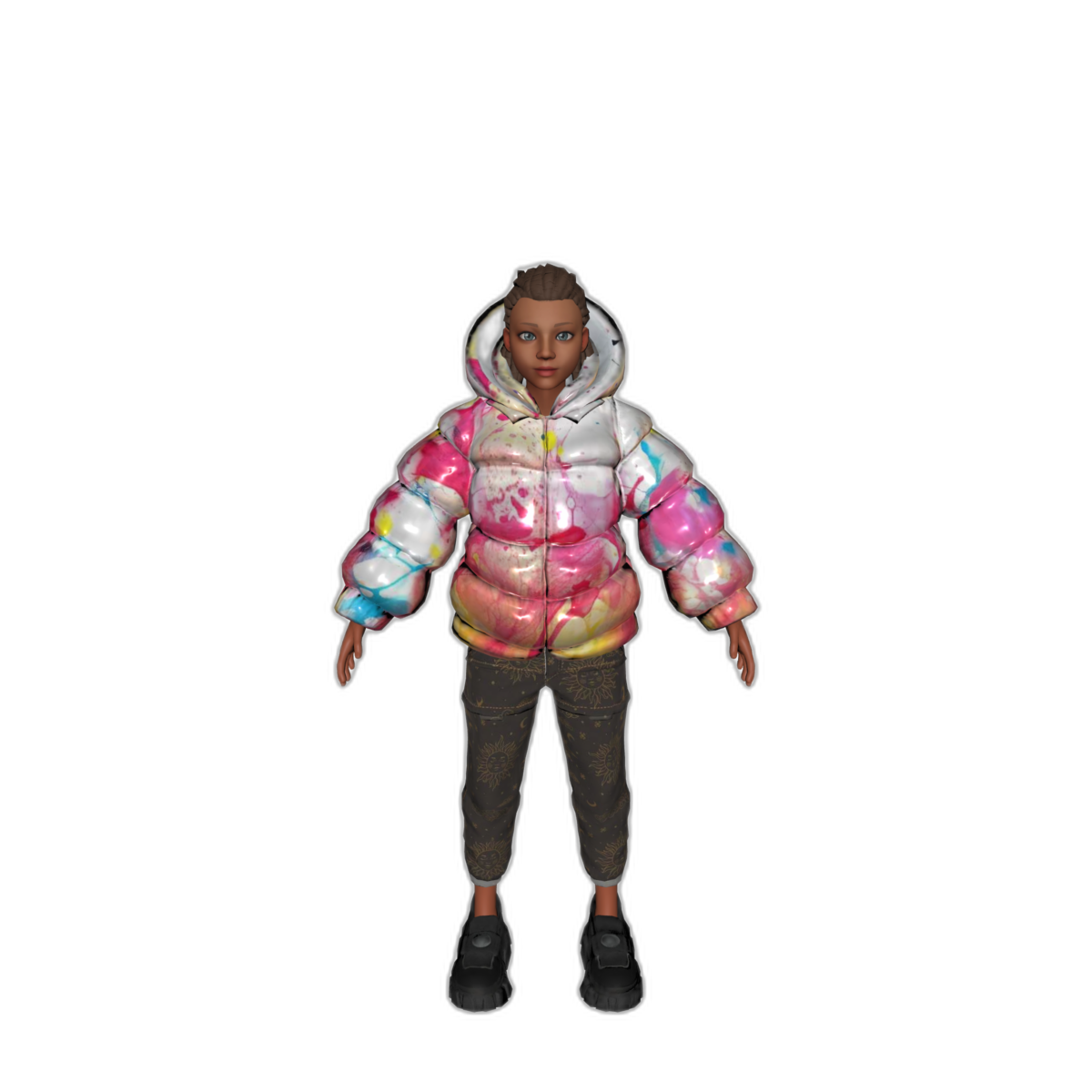}{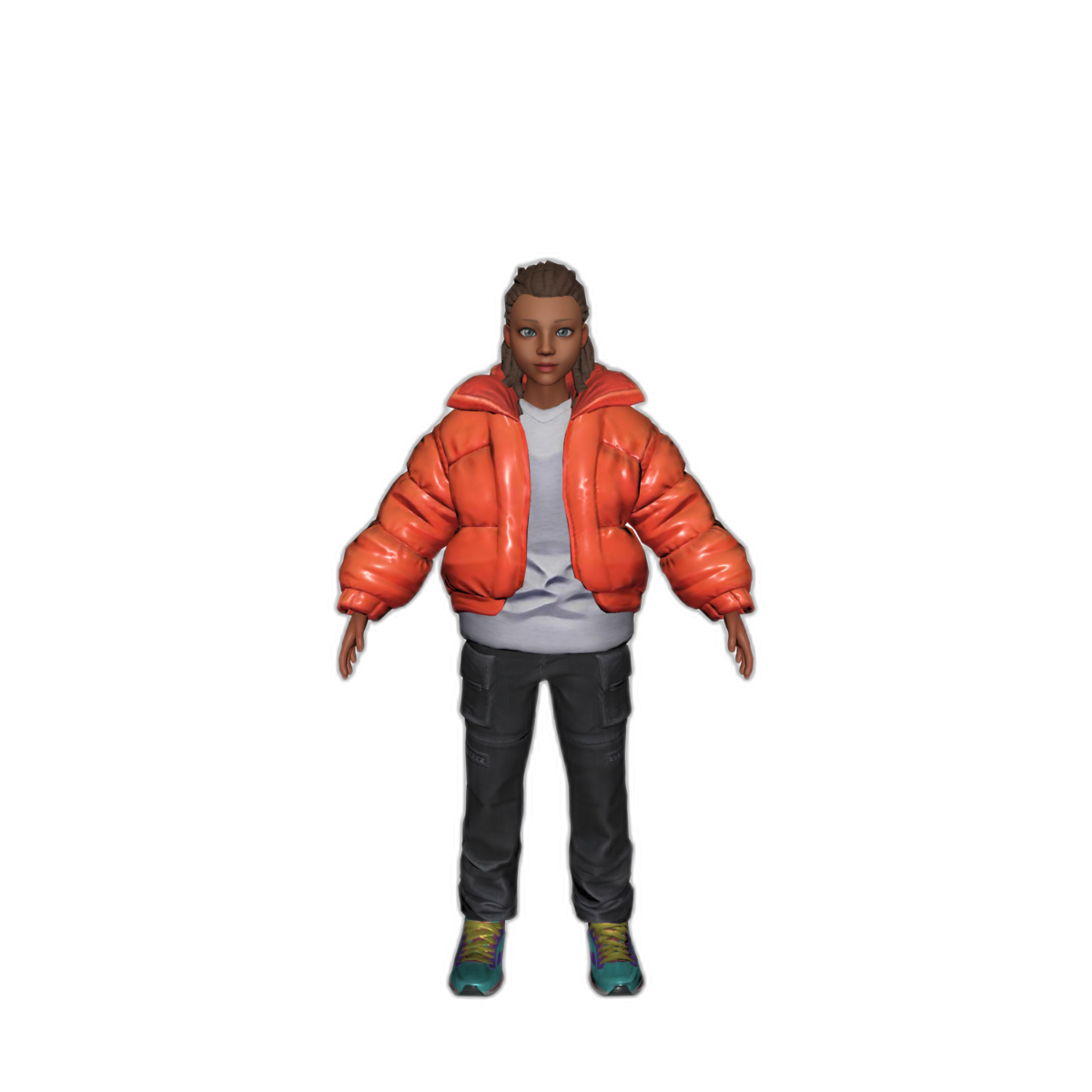}{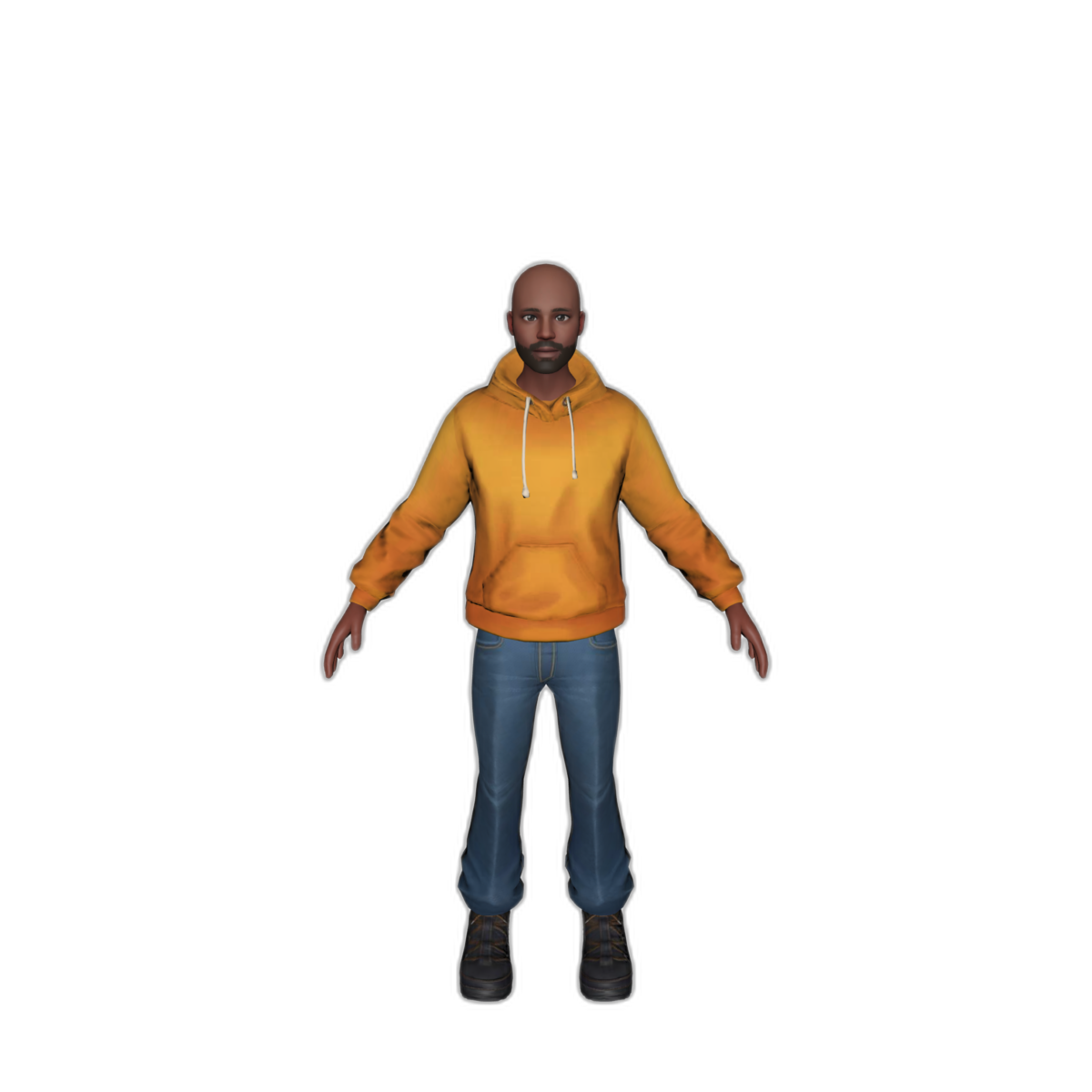}
            {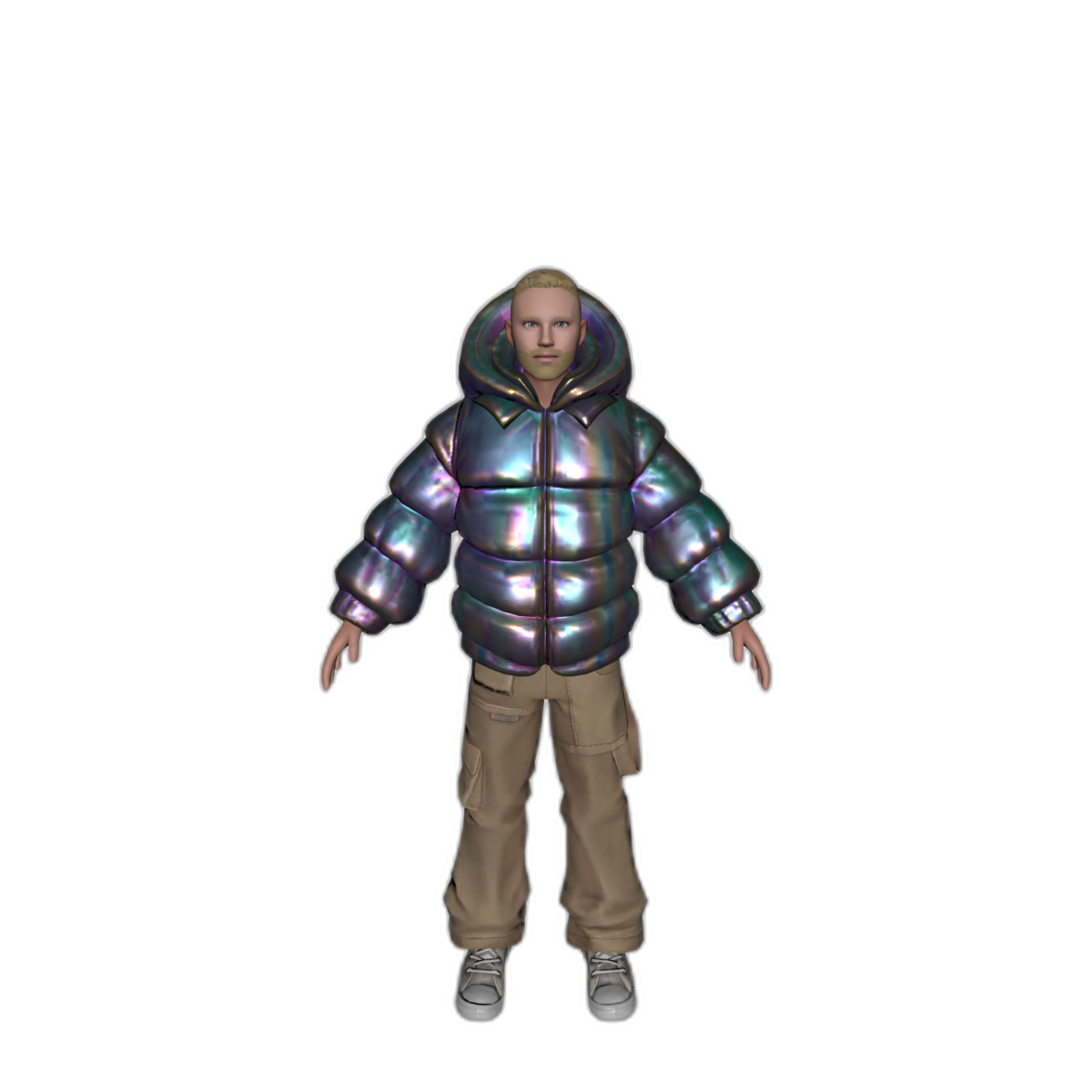}{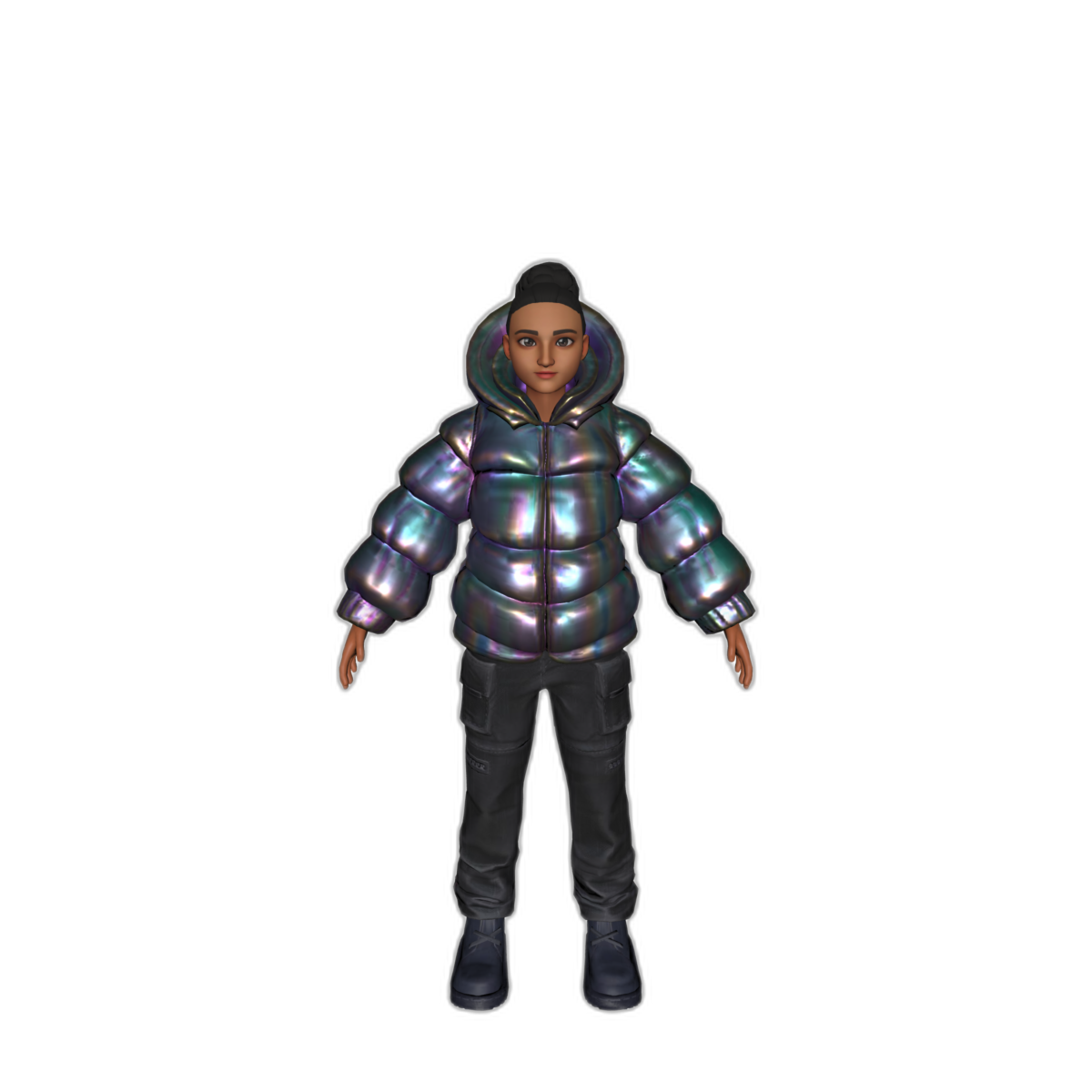}{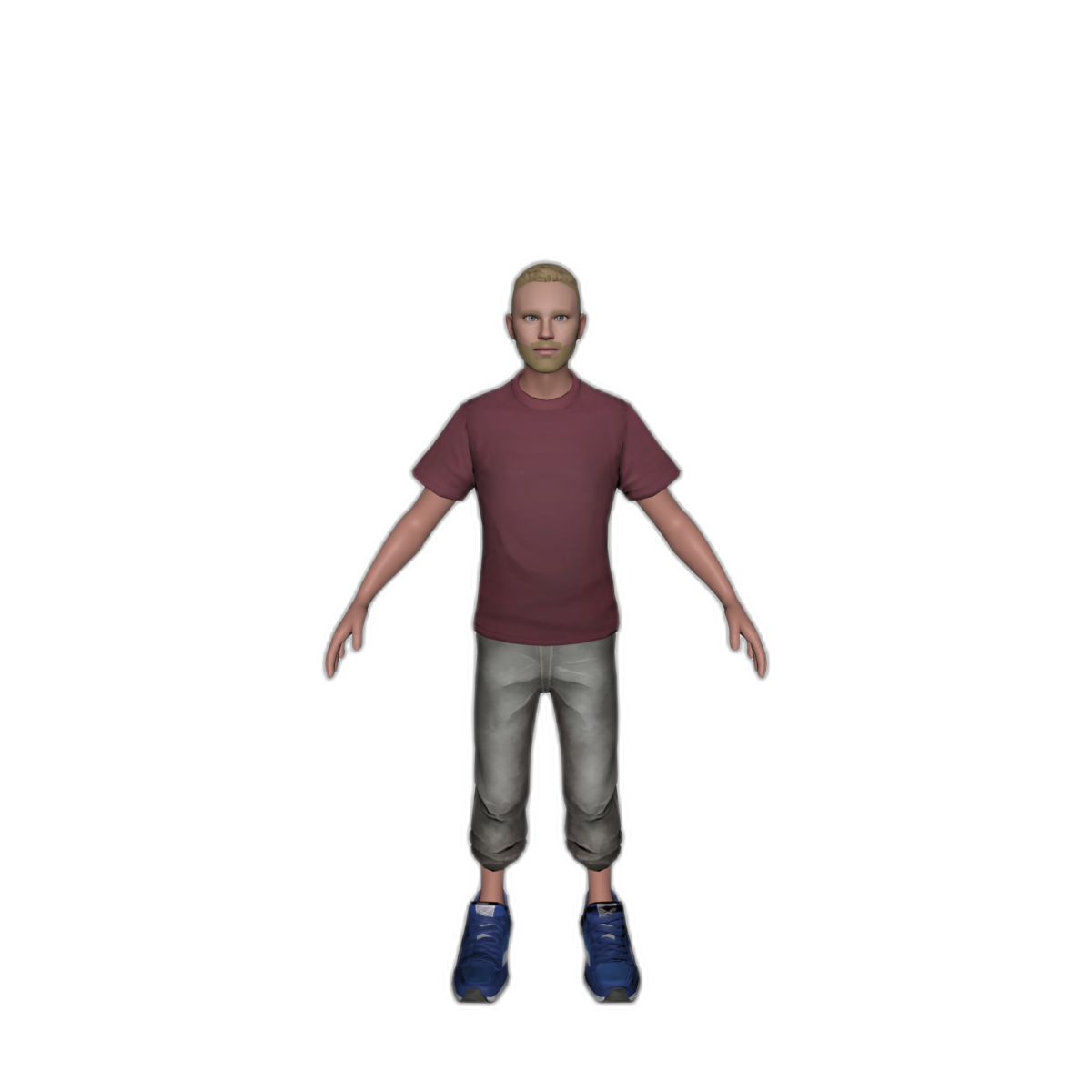}

        \qualrow{VLM}
            {5}{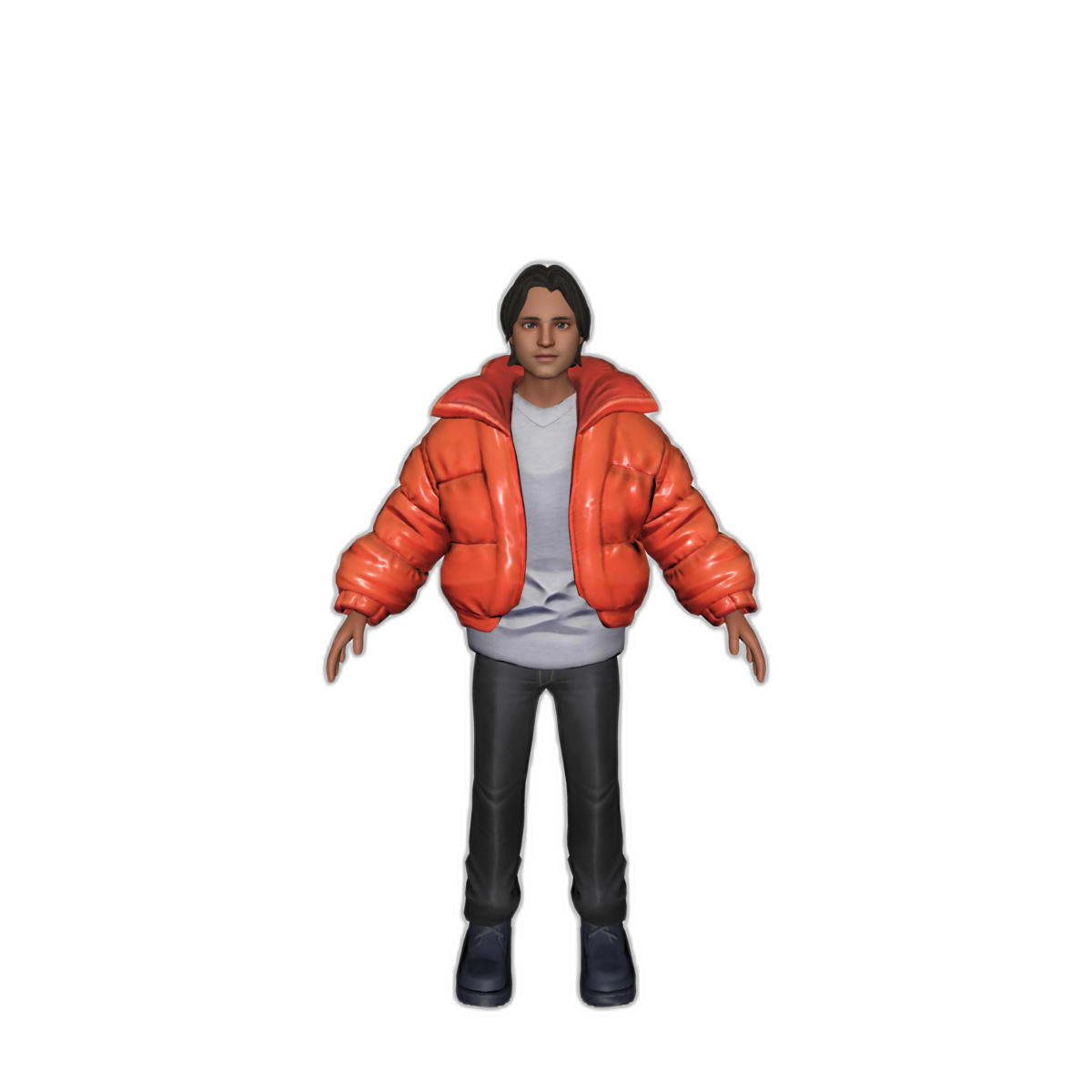}{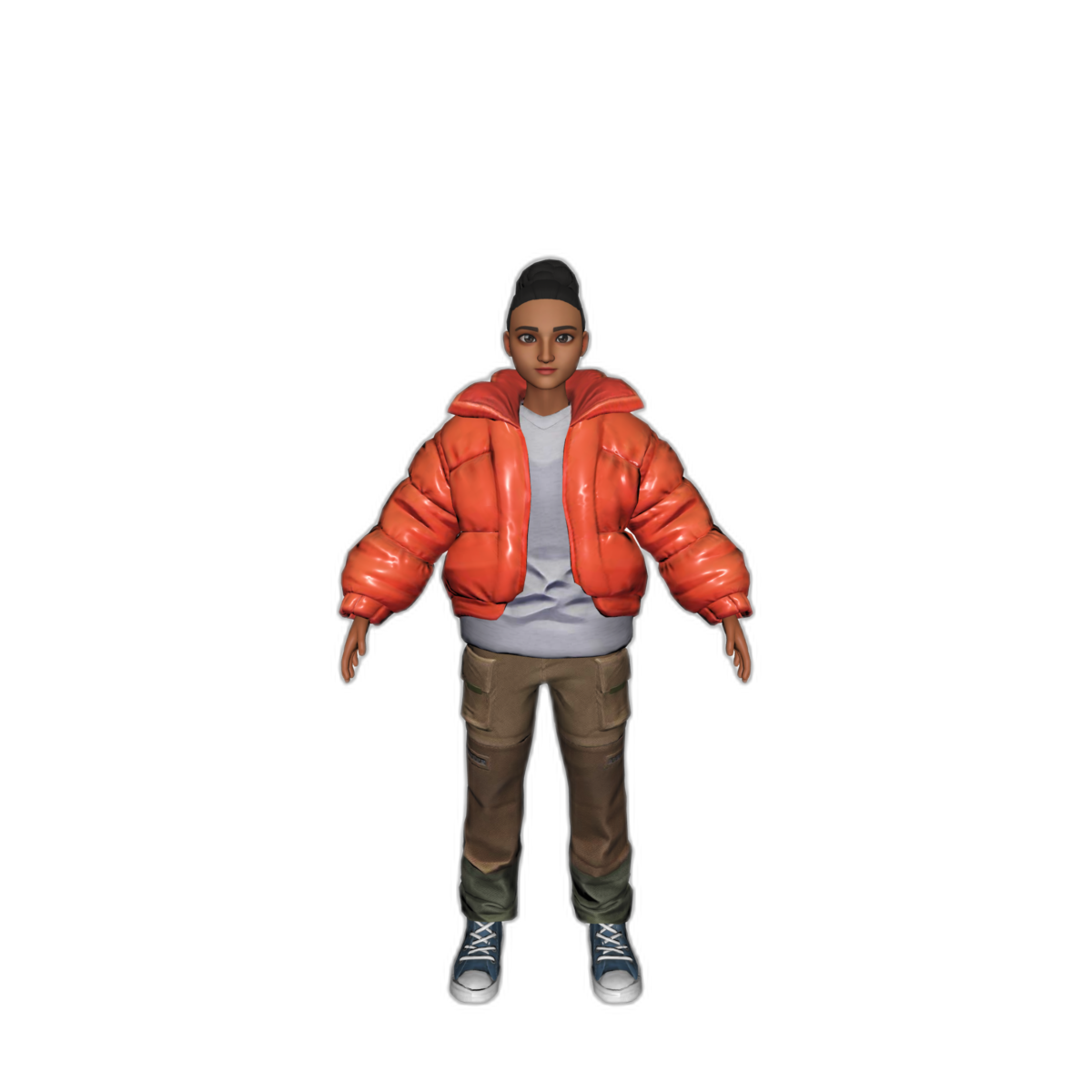}
            {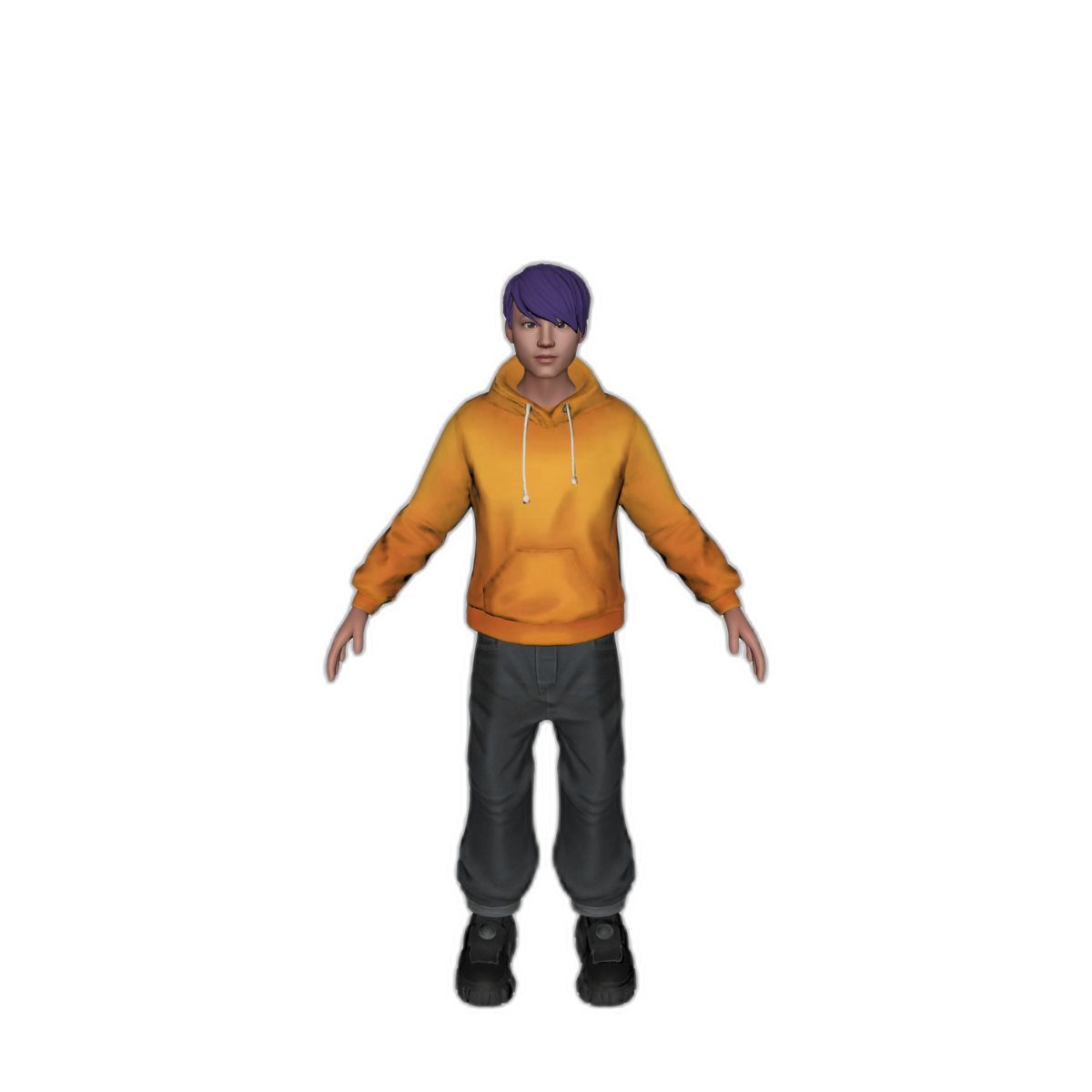}{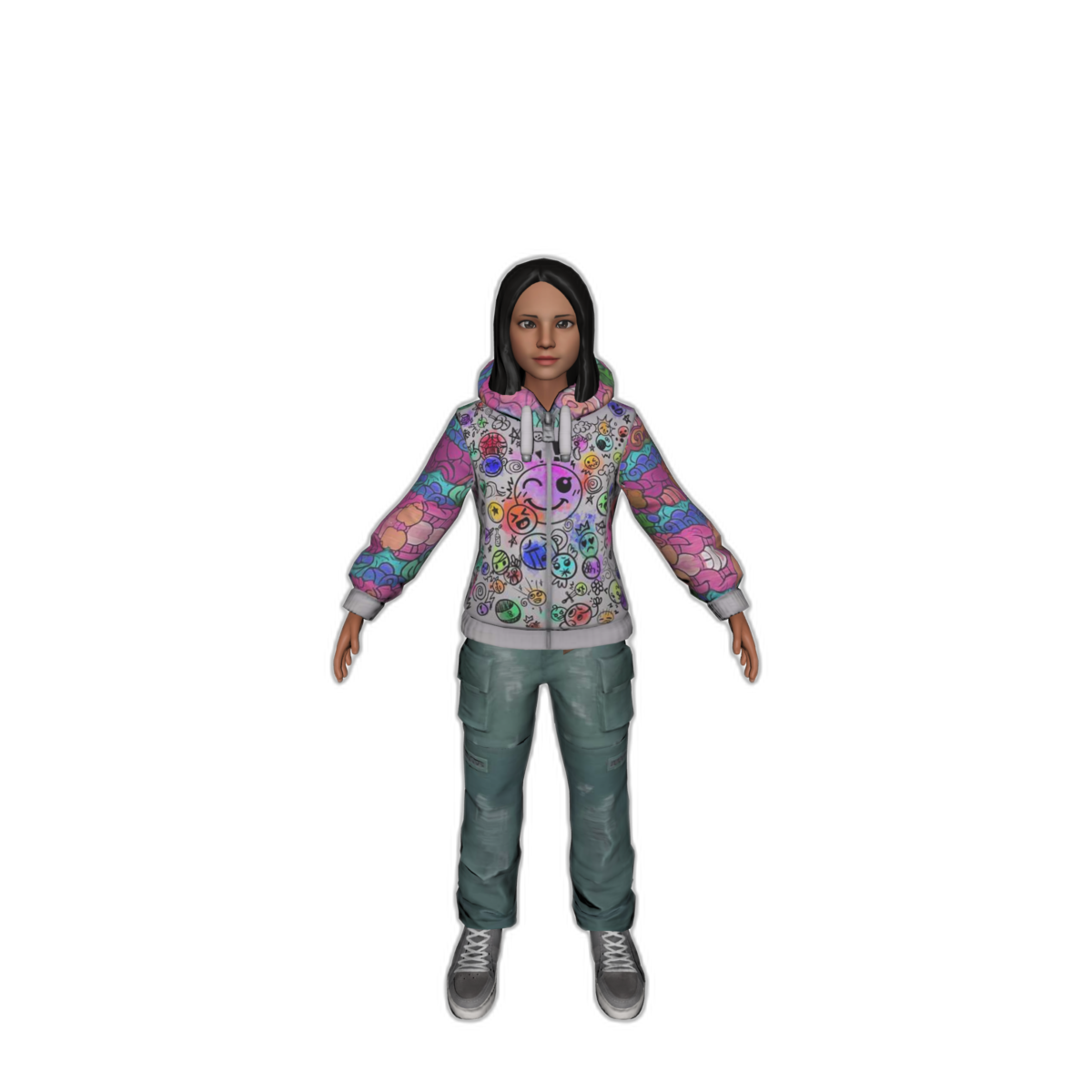}{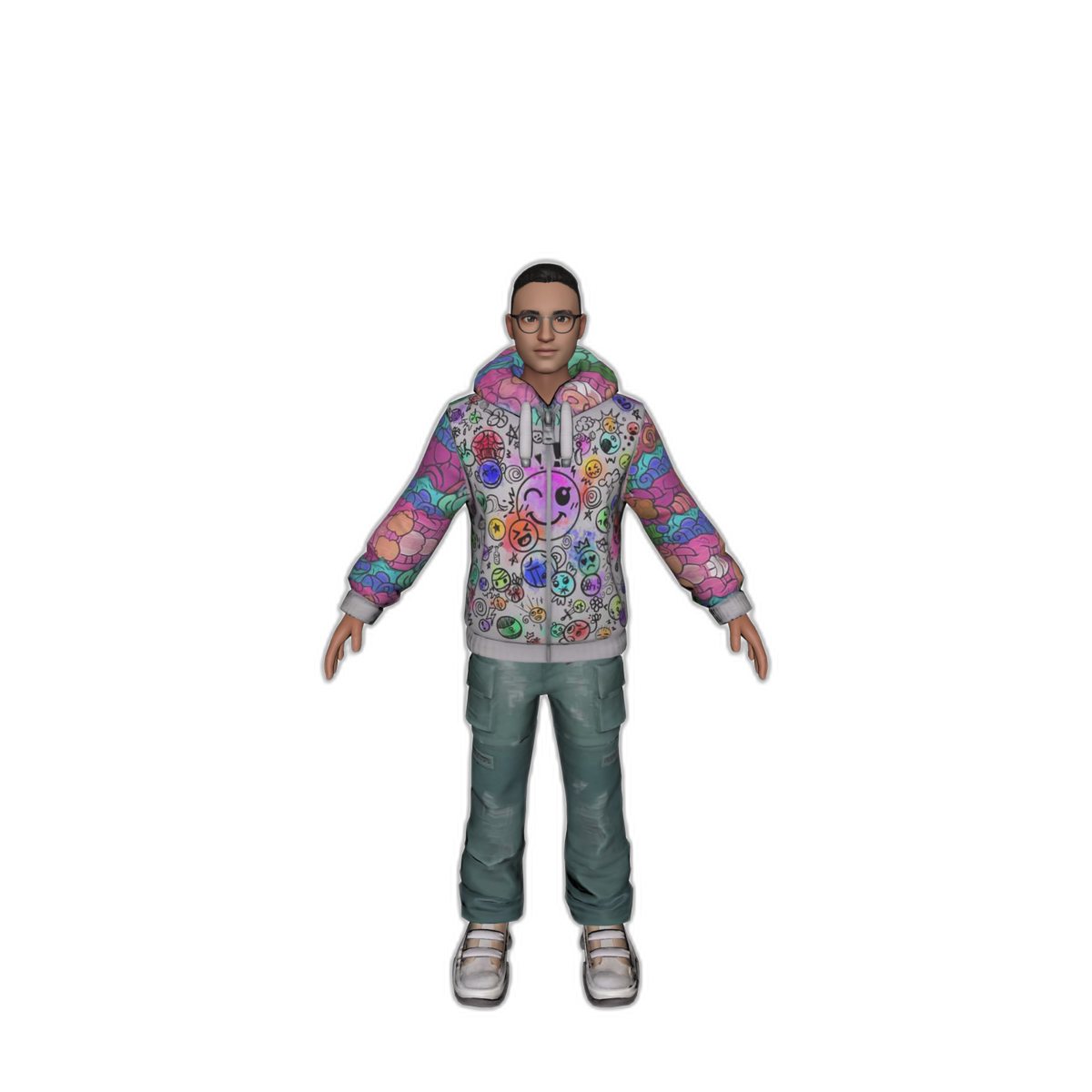}

        \qualrow{\textbf{GEAR}}
            {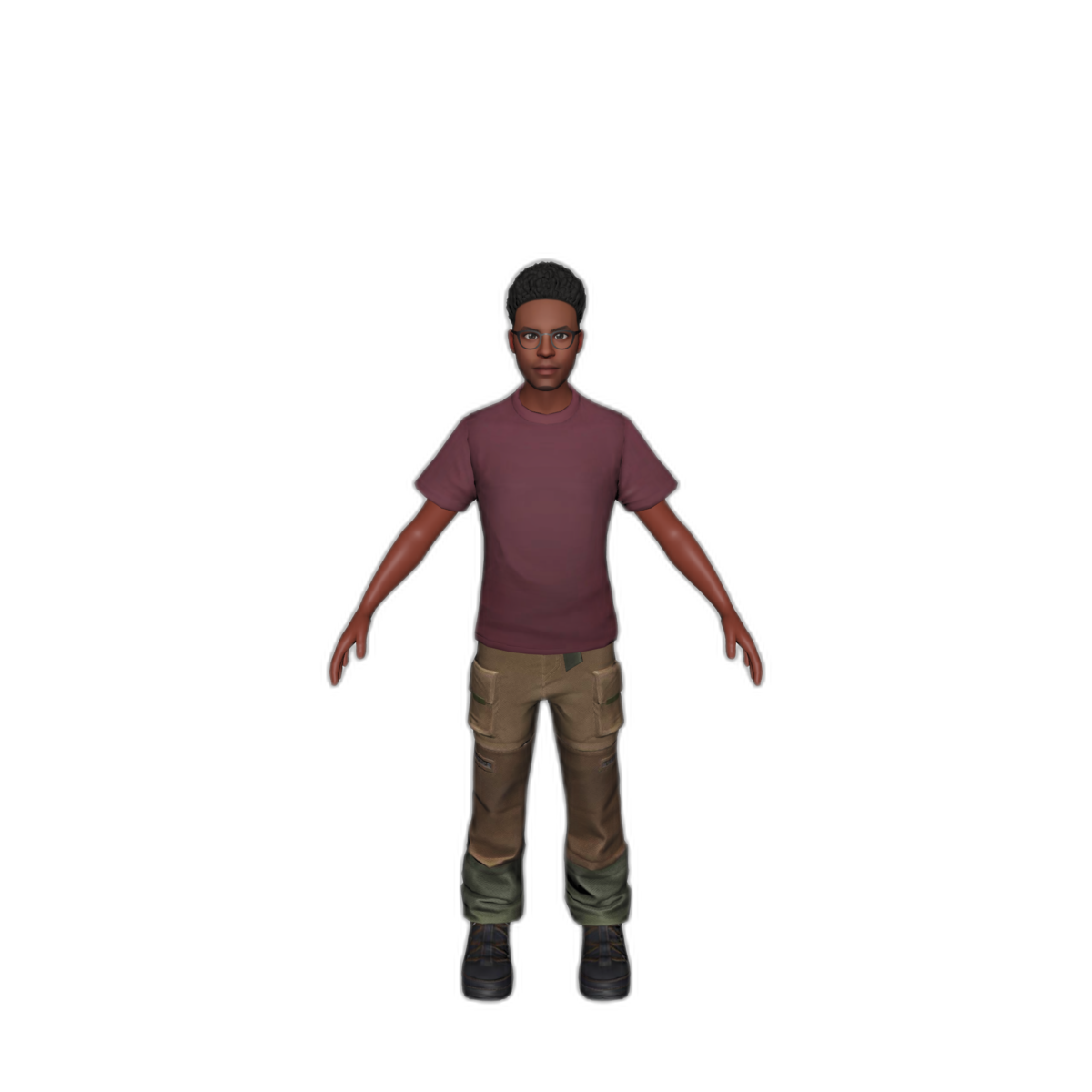}{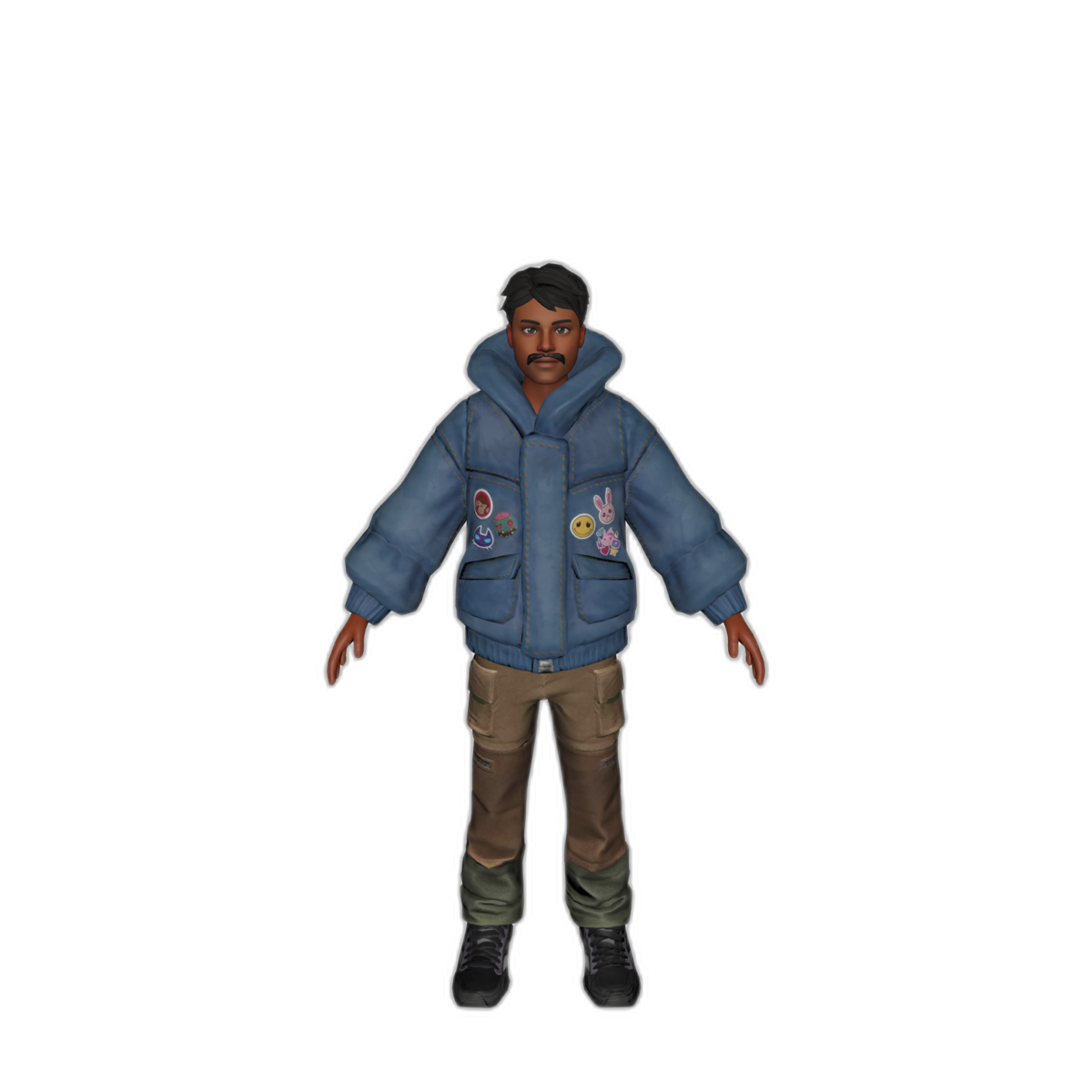}{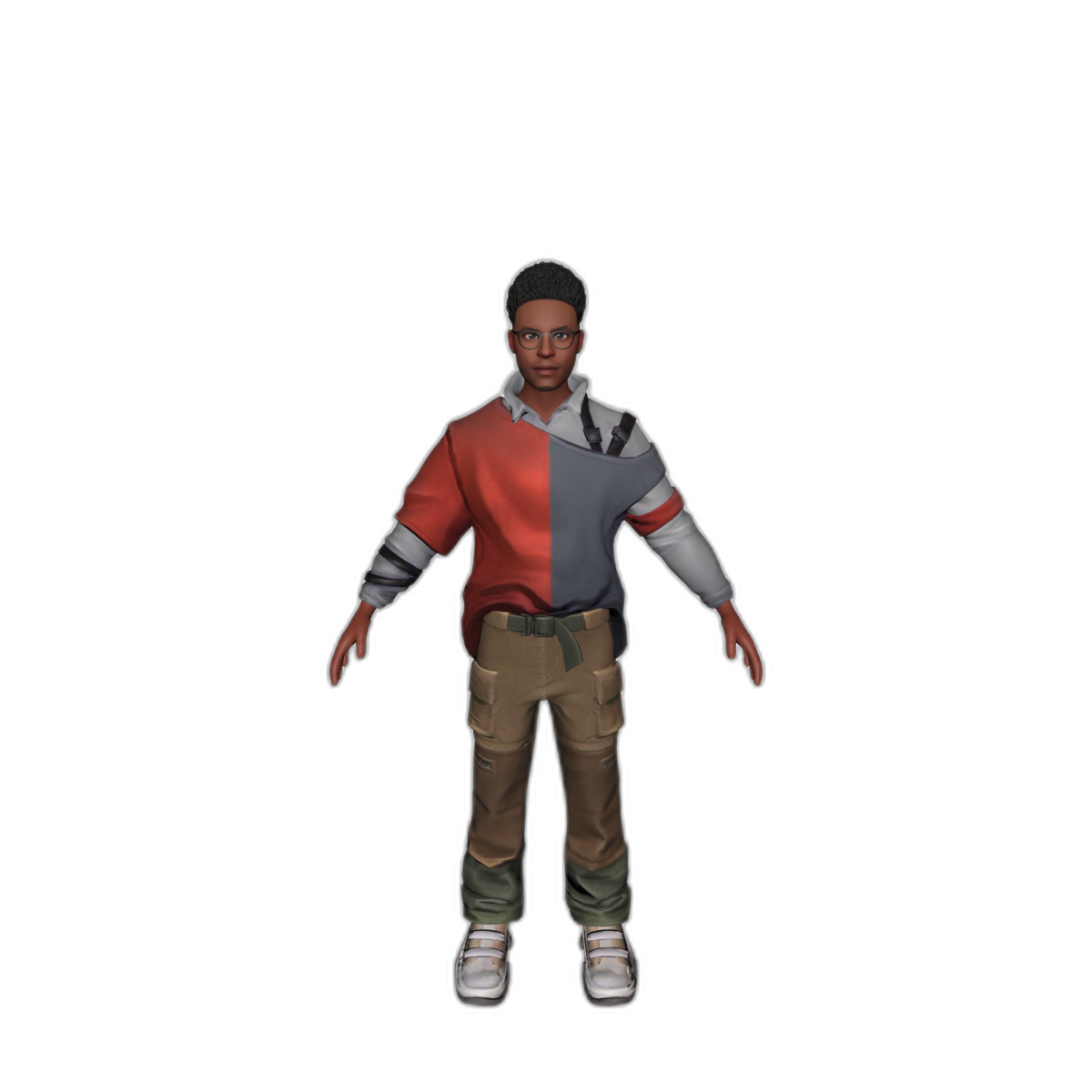}
            {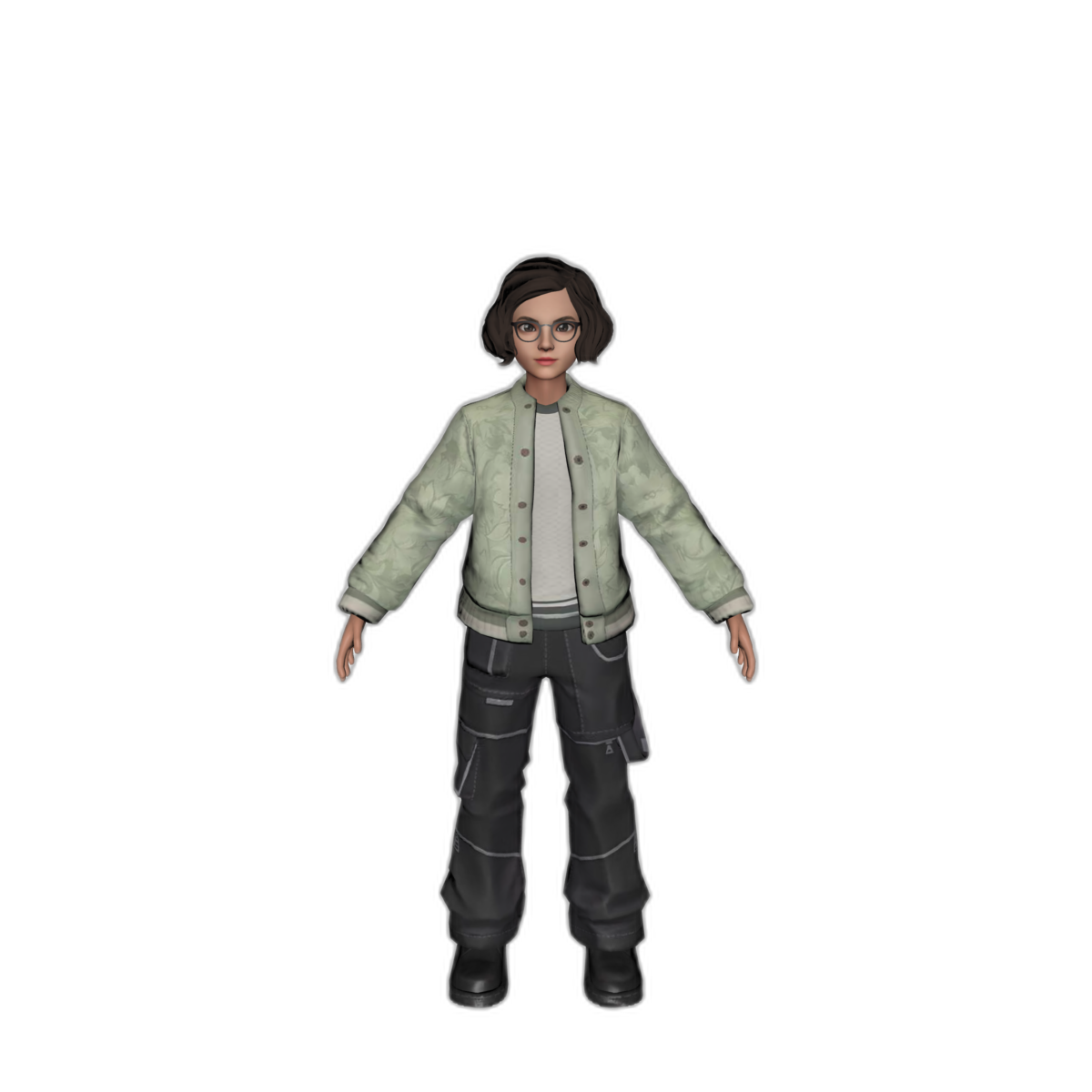}{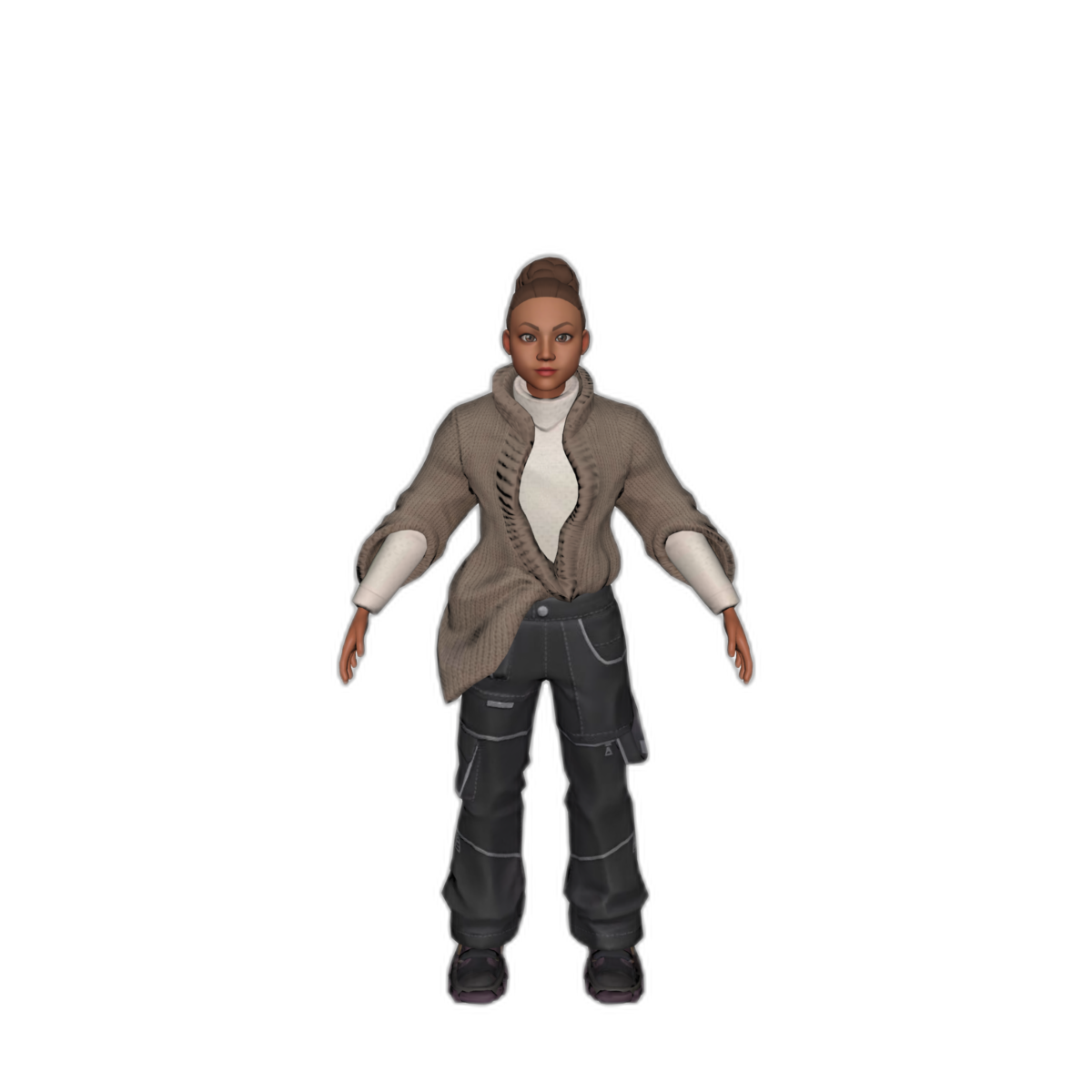}{12}

        \bottomrule

    \end{tabular}

    \caption{Qualitative comparison of the top-three
    retrieved outfits of 2 scenes. All methods retrieve from the same
    candidate pool. Outfits are shown in descending
    rank order.}

    \label{fig:qualitative_comparison}
\end{figure*}

\subsection{Qualitative Comparison}

Figure~\ref{fig:qualitative_comparison} illustrates the
top-three recommendations produced by the four methods
for two outdoor scenes.

Several notable behaviors emerge from these examples.
In the mountainous scene, Aesthetic Predictor retrieves
brightly colored and patterned outerwear that appears
less visually harmonious with the natural surroundings.
This illustrates a potential limitation of aesthetic-based
ranking, where visually appealing garments may not
necessarily suit the environmental context. VLM Retrieval produces three nearly identical orange
jackets in the mountainous scene and two visually
similar patterned outfits in the fantasy scene.
Although these recommendations may individually be
suitable, their redundancy reduces the variety of
options available to the user. CLIP ranks the same yellow outfit first in both
scenes, whereas GEAR returns different top-ranked
outfits, including a burgundy shirt for the mountainous
scene and a light-green jacket for the fantasy scene.
GEAR also retrieves visually varied alternatives
within each recommendation set. These examples illustrate potential differences in
contextual suitability, recommendation diversity,
and scene sensitivity.

\subsection{User Study}

\subsubsection{Design and Procedure}

We conducted a user study with ten participants aged 20--22 years
to assess the perceived visual appeal and contextual suitability
of retrieved outfits across 20 scenes. For each scene, participants
compared the top-three recommendations from four retrieval methods,
all operating on the same pool of 100 candidate outfits.

To minimize presentation bias, method identities were concealed
using randomized labels (A--D), and scene presentation order
was randomized for each participant. Participants could preview
the recommended outfits against the corresponding scene before
selecting their preferred recommendation set independently for
\emph{Visual Appeal} and \emph{Context Fit}. Overlapping
recommendations across methods were permitted.

\subsubsection{Analysis}

For each evaluation criterion, we obtained 200 selections
across 10 participants and 20 scenes. We compared GEAR
against each baseline using paired differences in
participant-level selection rates. Statistical significance
was assessed using two-sided exact sign-flip permutation
tests, enumerating all $2^{10}$ possible sign assignments.
We report mean differences in percentage points, nominal
95\% confidence intervals based on Student's $t$
distribution, and paired effect sizes ($d_z$).

To account for multiple comparisons, we applied Holm
correction separately to the three GEAR--baseline
comparisons within each evaluation criterion.

\subsubsection{Results}

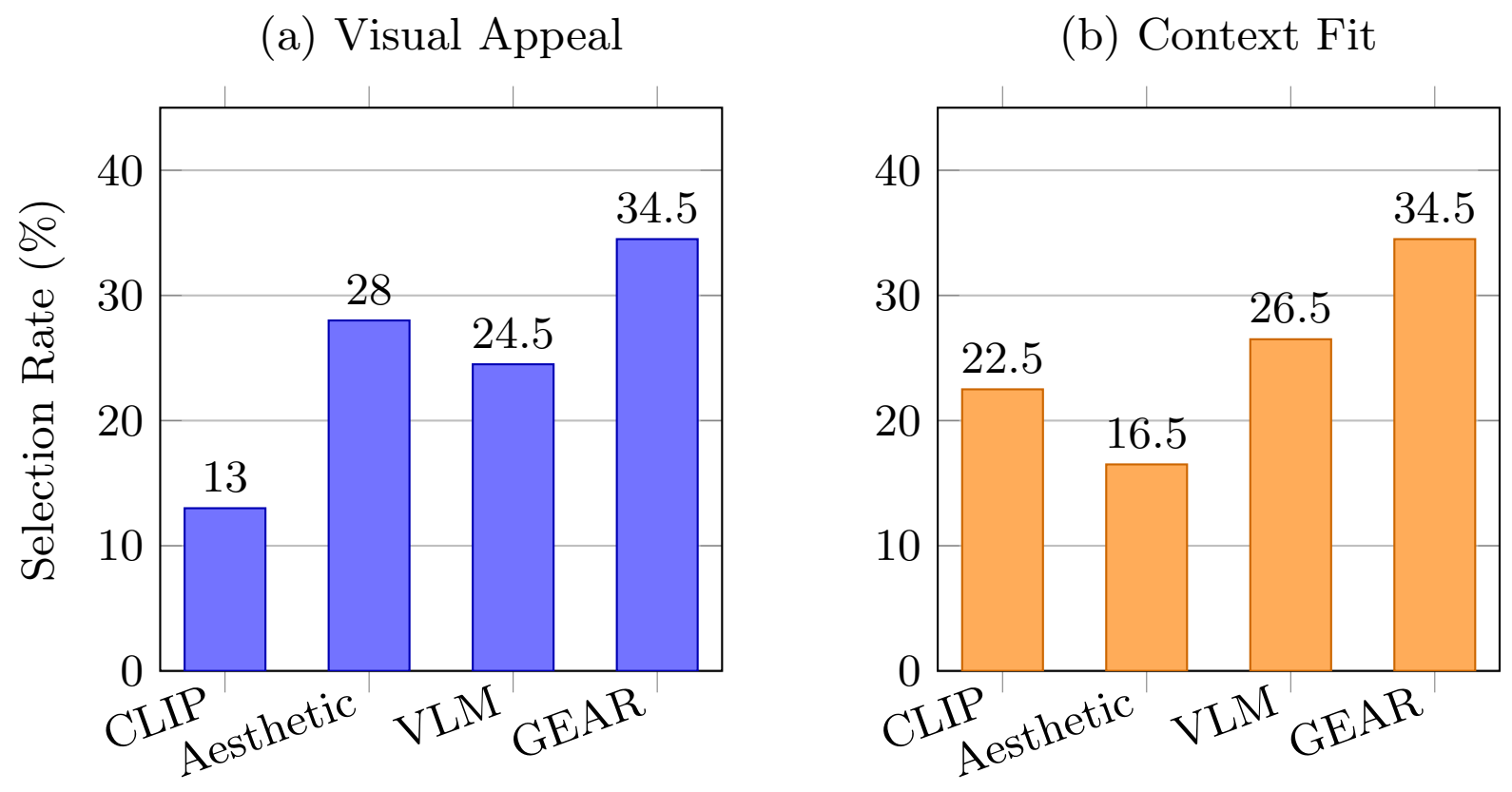
\begin{figure*}[t]
    \centering
    \begin{tikzpicture}
        \begin{groupplot}[
            group style={
                group size=2 by 1,
                horizontal sep=1.5cm
            },
            ybar,
            /pgf/bar width=16pt,
            width=0.45\textwidth,
            height=5.5cm,
            ymin=0,
            ymax=45,
            symbolic x coords={
                CLIP,
                Aesthetic,
                VLM,
                GEAR
            },
            xtick=data,
            xticklabels={
                CLIP,
                Aesthetic,
                VLM,
                GEAR
            },
            x tick label style={
                rotate=20,
                anchor=east,
                font=\small
            },
            ytick={0,10,20,30,40},
            nodes near coords,
            nodes near coords style={
                font=\small,
                /pgf/number format/fixed,
                /pgf/number format/precision=1
            },
            enlarge x limits=0.15,
            ymajorgrids=true,]

        \nextgroupplot[
            title={(a) Visual Appeal},
            ylabel={Selection Rate (\%)}
        ]

        \addplot[
            fill=blue!55,
            draw=blue!70!black
        ] coordinates {
            (CLIP,13.0)
            (Aesthetic,28.0)
            (VLM,24.5)
            (GEAR,34.5)
        };

        \nextgroupplot[
            title={(b) Context Fit}
        ]

        \addplot[
            fill=orange!65,
            draw=orange!80!black
        ] coordinates {
            (CLIP,22.5)
            (Aesthetic,16.5)
            (VLM,26.5)
            (GEAR,34.5)
        };

        \end{groupplot}
    \end{tikzpicture}

    \caption{User-study selection rates for the four
    retrieval methods across 20 travel scenes.
    Participants selected their preferred three-outfit
    set separately for (a) visual appeal and
    (b) context fit. Each criterion received 200
    scene-level selections from 10 participants.
    Percentages are computed separately for
    each criterion.}

    \label{fig:user_study_results}
\end{figure*}

GEAR received the largest observed share of selections for both criteria
(69/200, 34.5\%; Figure~\ref{fig:user_study_results}). Its mean differences
relative to VLM Retrieval were +10.0 percentage points for visual appeal and
+8.0 points for context fit, with both nominal CIs including zero
(Table~\ref{tab:user_study_comparisons}). The GEAR--CLIP visual-appeal and
GEAR--Aesthetic Predictor context-fit comparisons had unadjusted $p<0.05$ and
nominal CIs excluding zero, but no comparison survived Holm correction.

Preferences varied across participants: GEAR--VLM differences ranged from
$-25$ to $+55$ percentage points for visual appeal and from $-10$ to $+35$
for context fit. These results describe preferences for static
three-outfit sets; they do not establish a statistically reliable advantage
for GEAR or assess interactive VR use.

\begin{table}[t]
    \centering
    \small
    \setlength{\tabcolsep}{3pt}
    \renewcommand{\arraystretch}{0.9}
    \caption{Differences in selection rates between GEAR and the baseline methods across 10
    participants. $\Delta$ is the mean paired difference in percentage points
    (pp); positive values favor GEAR. CIs are nominal 95\% participant-level
    $t$ intervals, and $d_z$ is the paired standardized effect. Two-sided
    sign-flip $p$ values are shown before and after Holm adjustment within
    each criterion.}
    \label{tab:user_study_comparisons}
    \begin{tabular}{lrrrrr}
        \toprule
        Baseline & $\Delta$ (pp) & 95\% CI (pp) & $d_z$ & $p$ & Holm $p$ \\
        \midrule
        \multicolumn{6}{l}{\textit{Visual Appeal}} \\
        VLM & $+10.0$ & $[-11.0,31.0]$ & 0.34 & 0.332 & 0.664 \\
        CLIP & $+21.5$ & $[3.0,40.0]$ & 0.83 & 0.029 & 0.088 \\
        Aesthetic Predictor & $+6.5$ & $[-18.3,31.3]$ & 0.19 & 0.590 & 0.664 \\
        \midrule
        \multicolumn{6}{l}{\textit{Context Fit}} \\
        VLM & $+8.0$ & $[-2.8,18.8]$ & 0.53 & 0.172 & 0.188 \\
        CLIP & $+12.0$ & $[-1.5,25.5]$ & 0.64 & 0.094 & 0.188 \\
        Aesthetic Predictor & $+18.0$ & $[4.6,31.4]$ & 0.96 & 0.023 & 0.070 \\
        \bottomrule
    \end{tabular}
\end{table}

\section{Discussion}

\subsection{Analysis}

The automated and human evaluations capture complementary
aspects of retrieval quality. While the VLM baseline achieves
higher pairwise agreement, GEAR obtains slightly higher
mean Elo@5 and top-10 overlap, suggesting differences
between pairwise agreement and top-ranked recommendation
quality. The human study additionally shows higher
observed selection rates for GEAR in both visual appeal
and context fit. However, the differences are small in
the automated evaluation, and none of the human-study
comparisons remain statistically significant after
multiple-comparison correction. These findings do not
establish a reliable performance advantage for GEAR.

\subsection{Limitations}


The evaluation compares complete pipelines with different
query construction, encoders, and reranking stages, so it does
not isolate the contribution of image generation. A matched
text-only baseline using the same outfit description and aligned
encoder checkpoint remains unevaluated. Sensitivity to
reference-character appearance, framing, cropping, and generation
randomness has not been assessed. Direct comparisons with SAFRS
and GLOSS are also absent.

The benchmark covers only 10 scenes and 100 sampled candidates
from a larger collection. Its automated comparisons share scenes
and outfits, so their number should not be interpreted as
independent evidence across thousands of contexts. Seven Swiss
rounds expose each outfit to only seven opponents; the resulting
Elo order has not been tested for sensitivity to pairing or
update order. Mirrored disagreement is treated as a tie,
conflating uncertain judgments with genuine indifference.
Judge repeatability and agreement with independent human pairwise
judgments remain unvalidated.

Additionally, 
the qualitative examples illustrate potential differences in
contextual suitability and recommendation diversity, but cover
only two selected scenes. These observations may not generalize
across the benchmark, and recommendation diversity was not
systematically quantified.

The user study involves only 10 participants aged 20--22 years
evaluating static top-three recommendation sets across 20 scenes.
Its findings may not generalize to other populations, environments,
or candidate collections. Moreover, participants evaluated the
same scenes, so the reported statistical inference is conditional
on this scene collection. Set-level preferences provide evidence
about perceived visual appeal and contextual suitability but do
not independently validate the automated Elo rankings. The study does not evaluate in-engine usability or asset fitting quality.
Broader evaluations of both retrieval quality and interactive
performance are therefore needed.

\section{Conclusion}

We presented GEAR, a proposed retrieval method within a scene-conditioned
outfit system for interactive 3D environments. GEAR renders a VLM's outfit
suggestion on a reference character and uses the clothing crop to retrieve
existing assets.

On a benchmark of 10 scenes and 100 outfits, VLM achieves
the highest pairwise agreement, the primary metric, while GEAR
obtains slightly higher mean Elo@5 and top-10 overlap.
GEAR receives the largest observed share of user-study preferences
for both visual appeal and contextual suitability, although no
GEAR--baseline comparison remains statistically significant after
Holm correction. These findings suggest competitive performance
within the evaluated setting. Further evaluation should isolate
the contribution of image generation and assess robustness and
interactive performance.

\begin{credits}
\subsubsection{\ackname}
This research is funded by Vietnam National University - Ho Chi Minh City (VNU-HCM) under Grant Number C2026-18-12. Trung-Nghia Le was funded by the Postdoctoral Scholarship Programme of Vingroup Innovation Foundation (VINIF), VinUniversity, code VINIF.2025.STS.14.
\end{credits}

\bibliographystyle{splncs04}
\bibliography{ref}

\end{document}